\documentclass[10pt,twocolumn,letterpaper]{article}

\usepackage[pagenumbers]{cvpr} 

\definecolor{cvprblue}{rgb}{0.21,0.49,0.74}
\usepackage[pagebackref,breaklinks,colorlinks,allcolors=cvprblue]{hyperref}

\def\paperID{10207} 
\def\confName{CVPR}
\def\confYear{2025}

\usepackage{makecell}
\usepackage[numbers]{natbib}

\usepackage{amsmath}

\usepackage[utf8]{inputenc} 
\usepackage[T1]{fontenc}    
\usepackage{hyperref}       
\usepackage{url}            
\usepackage{booktabs}       
\usepackage{amsfonts}       
\usepackage{nicefrac}       
\usepackage{microtype}      
\usepackage{xcolor}         

\usepackage{url}
\usepackage{times}
\usepackage{epsfig}
\usepackage{amsmath}
\usepackage{amssymb}
\usepackage{graphicx}
\usepackage{multirow}
\usepackage{multicol}
\usepackage{algorithmic}
\usepackage{booktabs}
\usepackage{enumitem}
\usepackage{colortbl}  
\usepackage{array}   
\usepackage{pifont}
\usepackage{xspace}
\definecolor{gray0}{gray}{0.9}

\newcommand{\Ours}{\textup{DynRefer}\xspace}

\title{\Ours: Delving into Region-level Multimodal Tasks via Dynamic Resolution}

\begin{document}

\author{Yuzhong Zhao\textsuperscript{1}\footnotemark[1]  \quad 
Feng Liu\textsuperscript{1}\footnotemark[1]  \quad
Yue Liu\textsuperscript{1}\quad
Mingxiang Liao\textsuperscript{1}\quad
Chen Gong\textsuperscript{2}\\
Qixiang Ye\textsuperscript{1}\quad
Fang Wan\textsuperscript{1$\dagger$}\\
\textsuperscript{1}University of Chinese Academy of Sciences
\quad
\textsuperscript{2}University of Virginia\\
{\tt\small zhaoyuzhong20@mails.ucas.ac.cn}\quad
{\tt\small liufeng20@mails.ucas.ac.cn}\\
{\tt\small liuyue171@mails.ucas.ac.cn}\quad
{\tt\small liaomingxiang20@mails.ucas.ac.cn}\\
{\tt\small chengong@virginia.edu}\quad
{\tt\small qxye@ucas.ac.cn}\quad
{\tt\small wanfang@ucas.ac.cn}
}

\twocolumn[{
\renewcommand\twocolumn[1][]{#1}%
\maketitle
\begin{center}
    \centering
  	\captionsetup{type=figure}
	\includegraphics[width=1\linewidth]{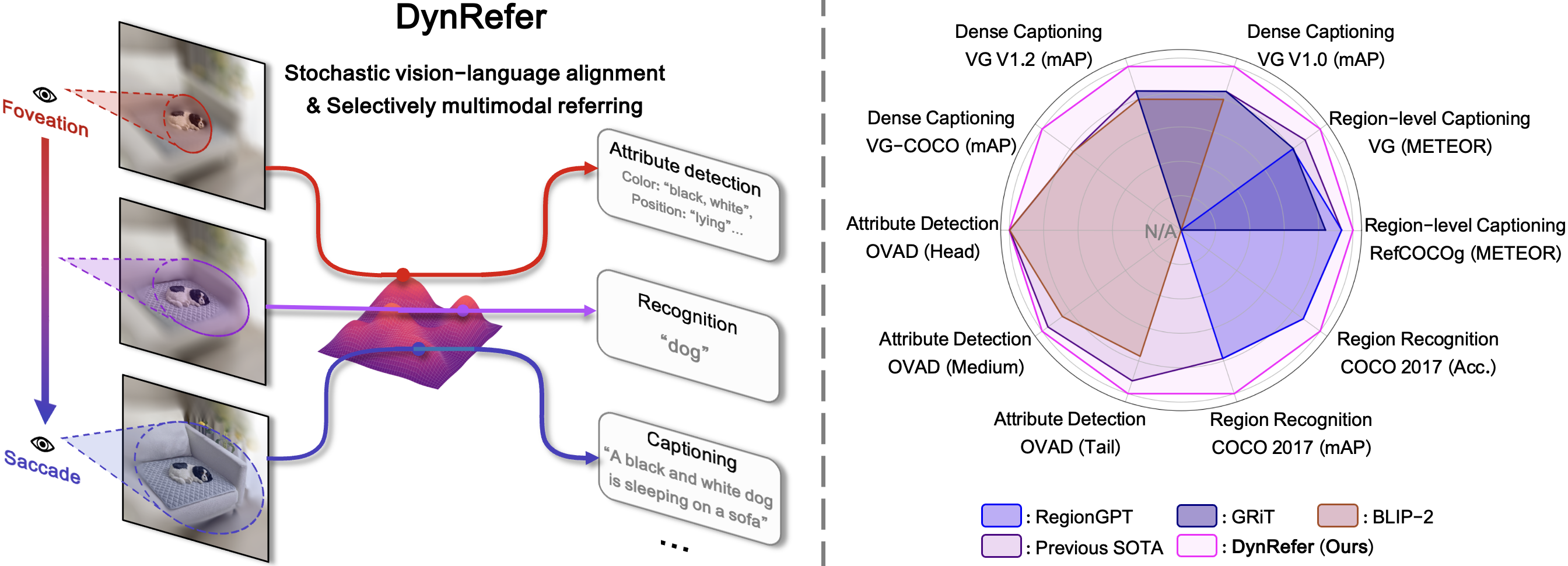}\\[-3mm]
    \vspace{1em}
 \captionof{figure}{
 \textbf{Left:} Illustration of our \Ours approach, which dynamically determines proper region views for each task through stochastic vision-language alignment and selectively multimodal referring.
 \textbf{Right:} Performance comparison on region-level multimodal tasks.
 }
	\label{fig:1}

\end{center}
}]

\renewcommand{\thefootnote}{\fnsymbol{footnote}}
\footnotetext[1]{~Equal contribution. $\dagger$ Corresponding Author.}

\maketitle
\begin{abstract}
One fundamental task of multimodal models is to translate referred image regions to human preferred language descriptions.
Existing methods, however, ignore the resolution adaptability needs of different tasks, which hinders them to find out precise language descriptions. In this study, we propose a DynRefer approach, to pursue high-accuracy region-level referring through mimicking the resolution adaptability of human visual cognition. 
During training, DynRefer stochastically aligns language descriptions of multimodal tasks with images of multiple resolutions, which are constructed by nesting a set of random views around the referred region. 
%
During inference, DynRefer performs selectively multimodal referring by sampling proper region representations for tasks from the nested views based on image and task priors.
This allows the visual information for referring to better match human preferences, thereby improving the representational adaptability of region-level multimodal models. 
Experiments show that DynRefer brings mutual improvement upon broad tasks including region-level captioning, open-vocabulary region recognition and attribute detection.
Furthermore, DynRefer achieves state-of-the-art results on multiple region-level multimodal tasks using a single model. 
Code is available at \url{https://github.com/callsys/DynRefer}.

\end{abstract}
\section{Introduction}
\label{sec:intro}

Region-level multimodal tasks, as a means of communicating referred information with computer, constitute an important branch of artificial intelligence~\cite{Johnson2016DenseCap, ovad, osprey, guo2024regiongpt}. 
These tasks involve translating specific image regions to language descriptions based on task requirements such as open-vocabulary region recognition~\cite{Radford2021CLIP}, attribute detection~\cite{ovad, ovarnet}, region-level captioning~\cite{osprey, Peng2023Kosmos2, Chen2023Shikra, guo2024regiongpt}.
Existing methods~\cite{Johnson2016DenseCap,Wu2022GRIT,ovad,ovarnet,regionclip} using image regions under fixed resolution as inputs remain lacking adaptability to capture detailed region information or rich global context.

\begin{figure*}[t]
	\includegraphics[width=1.0\linewidth]{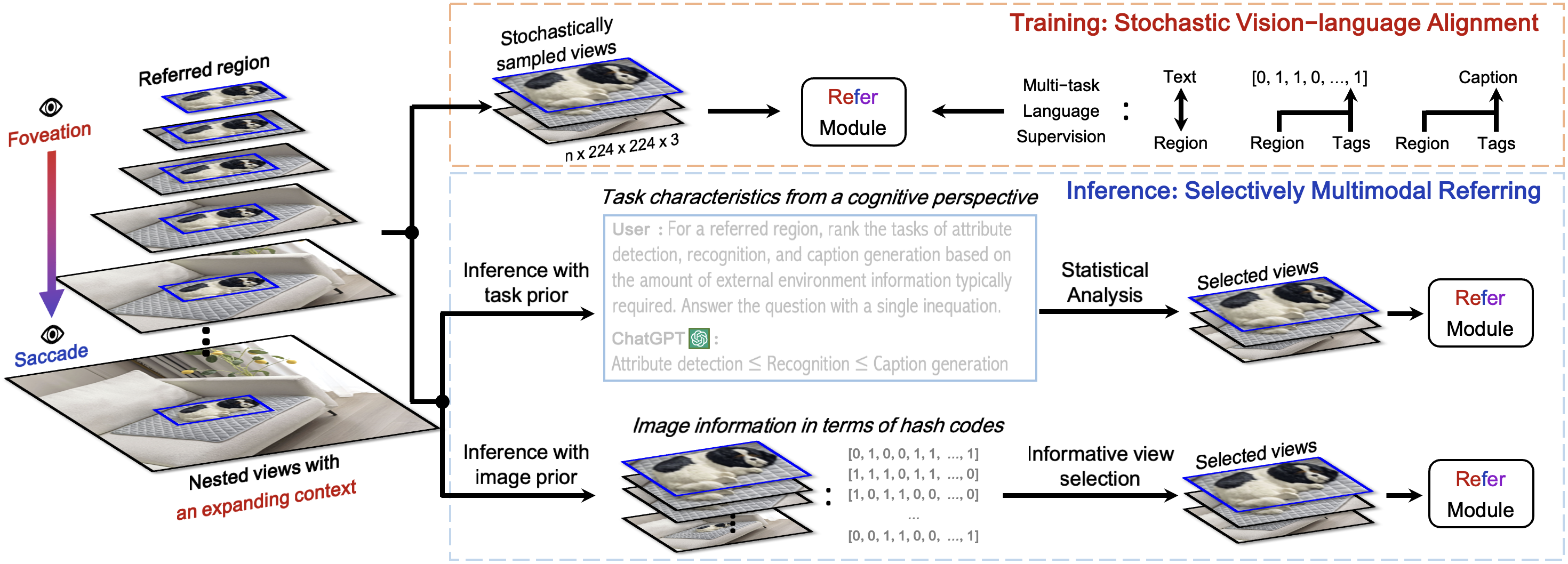}\\
    \caption{Diagram of the proposed \Ours. The ``dynamic" capability is achieved through a stochastic vision-language alignment procedure during training (upper) and a selectively multimodal referring procedure during reference (lower).
    During training, the input image is cropped and resized to multiple views surrounding the referred region. The views are then randomly sampled to simulate an image with stochastic resolution. The sampled views are used to train a Refer Module (upper). During inference, the views are sampled based on task and image priors to meet the task requirements and human preference (lower).}
    \label{fig:2}
\end{figure*}

A naive solution is to increase the resolution of the entire input image to enrich region representations with finer details and broader context.
This solution, however, introduces a substantial computational overhead, as popular vision foundation models~\cite{DBLP:conf/iclr/DosovitskiyB0WZ21, fang2023eva, openclip} have already been puzzled by the computational complexity, $e.g.$, $\mathcal{O}(n^2)$ $w.r.t.$ the length of input sequence). 
{Additionally, increasing the resolution of input images requires the algorithm to process more irrelevant regions, aggregating the challenge to distinguish useful contextual information from noise.}

{As a reference,} the human visual cognition system can adjust focus of attention through processes like foveation and saccadic eye movements~\cite{Topography1990,EyeMove2018}, which implies dynamically varying image resolution according to task requirements.
For example, Figure~\ref{fig:1}(left), when identifying attributes of a small region, humans tend to focus their gaze on that specific area ($i.e.$, foveation).
When aiming to provide a description of a region within its surrounding context, humans typically scan the area and its environment ($i.e.$, saccade).
Different from human perceptual capabilities, multimodal large language models (MLLMs)~\cite{Li2023BLIP2, guo2024regiongpt, osprey, pvit} treat all visual regions equally, which leads to poor encoding of the referred regions and contextual information, hampering model's adaptability to diverse tasks.

{Inspired by dynamic resolution characteristics of the visual cognition system,} we propose a {\textit{simple-yet-effective} computational approach, \Ours,} to address {the adaptability challenge} of region-level multimodal tasks, Fig.~\ref{fig:1}, from the following two perspectives.
(i) \textit{Non-uniformity.} The referred region is represented as a high resolution image, while irrelevant regions are either represented as a low resolution image or removed entirely. This forces the model to focus on query relevant regions, leading to better information encoding.
(ii) \textit{Adaptability.} The resolution of the image is dynamically adjusted $w.r.t.$  the specific language output required by the task. 
Adaptation enables the model to better align with human preferences. Specifically, it enhances the resolution of the referred region when fine details are required, and improves the resolution of the overall environment when a context-aware description is needed.

\Ours pursues high-accuracy region-level referring by performing stochastic vision-language alignment during training, and selectively multimodal referring during inference, Fig.~\ref{fig:2}.
For stochastic vision-language alignment, we create images with stochastic resolutions by combining randomly nested image views around the referred region. These images are then embedded and aligned with the desired language descriptions for {region-level multimodal} tasks. 
{
For selectively multimodal referring, we select appropriate image views to form a proper region representation based on task prior and image prior, Fig.~\ref{fig:2}(lower). When task types are known in advance, we select views based on the attributes and characteristics of the task. For example, for attribute detection that require fine details, we select contextless but detail-rich views. When task types are unknown, we select views based on image priors by maximizing the total information of the combined views using a greedy search algorithm.
This enables the model to generate task-specific outputs aligning with human preferences.
}%

Extensive experiments conducted on OVAD~\cite{ovad}, COCO~\cite{coco}, Visual Genome~\cite{krishna2017visual}, and RefCOCOg~\cite{yu2016modeling} show that \Ours enjoys high representational capacity and strong task adaptability, 
With a single model, \Ours is capable of executing multiple tasks and outperforms the state-of-the-art methods in open-vocabulary attribute detection, region recognition, and region-level captioning methods with significant margins, Fig.~\ref{fig:1}(right).
Specifically, \Ours respectively improves mAP by \textbf{1.1\%} on OVAD (Open-vocabulary attribute detection), accuracy by \textbf{8.8\%} on COCO (Open-vocabulary region recognition), mAP by \textbf{7.1\%} on Visual Genome V1.2 (Dense captioning), and CIDEr by \textbf{5.8} on RefCOCOg (Region-level captioning).

The contributions of this study are summarized as follows:
\begin{itemize}[leftmargin=*]
    \item We propose \Ours, a simple-yet-effective approach, to pursue high-accuracy region-level referring through mimicking the dynamic resolution mechanism of visual cognition.

     \item We design a stochastic vision-language alignment procedure to train dynamic resolution models, which constructs the implicit correspondence between dynamic resolution inputs and specific language outputs. We further propose a selectively multimodal referring procedure for dynamic resolution inference, which supports the adaptive prediction of language descriptions for referred regions.

    \item Experiments on multiple benchmarks show that 
    \Ours achieves the state-of-the-art results for multiple region-level multimodal tasks using a single model.
    
\end{itemize}

\section{Related Works}
\label{sec:relate_work}

\textbf{Vision-Language Models.} These methods aim to learn multimodal comprehension ability given image-text pairs.
Benefiting from powerful foundation models~\cite{vaswani2017attention, DBLP:conf/iclr/DosovitskiyB0WZ21, Devlin2018BERT, Zhang2022OPT, Chung2022Flan5} and huge amount of vision-language data corpus~\cite{schuhmann2022laion}, VLMs have achieved unprecedented performance across vision-language tasks such as semantic segmentation~\cite{zhao2023generative, wu2023datasetdm}, image-text retrieval~\cite{Li2022BLIP, Li2023BLIP2, Wang2022BEIT3, li2020oscar}, visual question answering (VQA)~\cite{Li2022BLIP, Li2023BLIP2, Dai2023InstructBLIP, Liu2023LLaVa} , image captioning~\cite{Li2022BLIP, Li2023BLIP2, Dai2023InstructBLIP, Liu2023LLaVa}, and few-shot learning~\cite{Alayrac2023Flamingo, yu2022coca}.
According to the training objectives, VLMs can be categorized into three types: (i) Image-text contrastive learning~\cite{Radford2021CLIP, openclip, regionclip, yu2022coca, Wang2022BEIT3}, (ii) Image-text matching~\cite{albef, Li2022BLIP, bao2022vlmo}, and (iii) Language modeling~\cite{Liu2023LLaVa, Li2022BLIP, Alayrac2023Flamingo, osprey, glamm}. 
To accomplish region-level tasks, some of these models~\cite{Peng2023Kosmos2, osprey, glamm, ovad, regionclip, wang2023allseeing, you2023ferret, omgllava, captionanything, segmentandcaption, tokenizeanything, dwibedi2024flexcap} are trained on region-text pairs to unlock their region-level comprehension ability. 

\textbf{Region-level Multimodal Tasks.} 
The acquisition of preferred semantics ($e.g.$, categories, attributes, captions) for given (referred) image regions is crucial for many multimodal tasks:
(i) Region recognition. With the rapid development of VLMs, classifying regions in an open set has become a common practice. The methods based on contrastive learning~\cite{regionclip, Long2023CapDet, Radford2021CLIP, openclip} get the class by calculating the similarity between region embeddings and text embeddings. While the methods based on language modeling~\cite{guo2024regiongpt, Liu2023LLaVa, Chen2023Shikra, pvit, zhang2023gpt4roi} query the large language model (LLM) to select the most likely class of given regions among an open set. 
(ii) Attribute detection. With the release of large-scale attribute datasets including COCO Attributes~\cite{cocoattributes}, Visual Genome~\cite{krishna2017visual}, and VAW~\cite{pham2021learning}, recent studies~\cite{pham2021learning, yun2022attributes} realize attribute detection by training multi-class classification networks. Inspired by CLIP~\cite{Radford2021CLIP}, OVAD~\cite{ovad}, OvarNet~\cite{ovarnet} learn to predict attributes from captions, which rely less on densely annotated attributes and can make predictions in an open vocabulary manner.
(iii) Region-level captioning. The generation of region-level captions based on large multimodal models (LMMs) has become a widespread practice~\cite{Chen2023Shikra, Peng2023Kosmos2, glamm, osprey, sun2023alpha}. GRiT~\cite{Wu2022GRIT} unifies the training of classification and captioning by treating object categories as brief captions. CapDet~\cite{Long2023CapDet} and DetCLIPv3~\cite{detclipv3} further combine dense captioning with open-world detection in a pretraining setup.

The trend of exploiting region-level information for fine-grained vision-language tasks urges the development of resolution adaptability, which is crucial to improve the accuracy of recognition, attribute detection, and region-level captioning by dynamically using the context information. 
Furthermore, for the multiple types of referring tasks, existing methods ignore the inherent similarity between region-level multimodal tasks.
There is an urgent requirement to unify these tasks from the perspective of model training. 
Such unification is expected to bring mutual improvement among tasks so that state-of-the-art results can be achieved for all tasks with a single model.

\textbf{Dynamic Resolution of Visual Cognition.} The research in the visual cognition area has shown that the human vision system has the capability of dynamic resolution. The fovea, situated in the central part of the retina, possesses the highest resolution view, while other parts of the retina dynamically perceive context views for details~\cite{Topography1990}. Recent research~\cite{EyeMove2018} has demonstrated that foveal and peripheral vision are closely linked and differences in appearance between peripheral and foveal vision can be adjusted through re-calibration~\cite{Attention2018}. In contrast, computer vision systems lack such a dynamic mechanism and instead capture only a static view~\cite{Peripheral2020}. To simulate the dynamic resolution mechanism through computer vision is non-trivial.

\begin{figure*}[t]
	\includegraphics[width=1.0\linewidth]{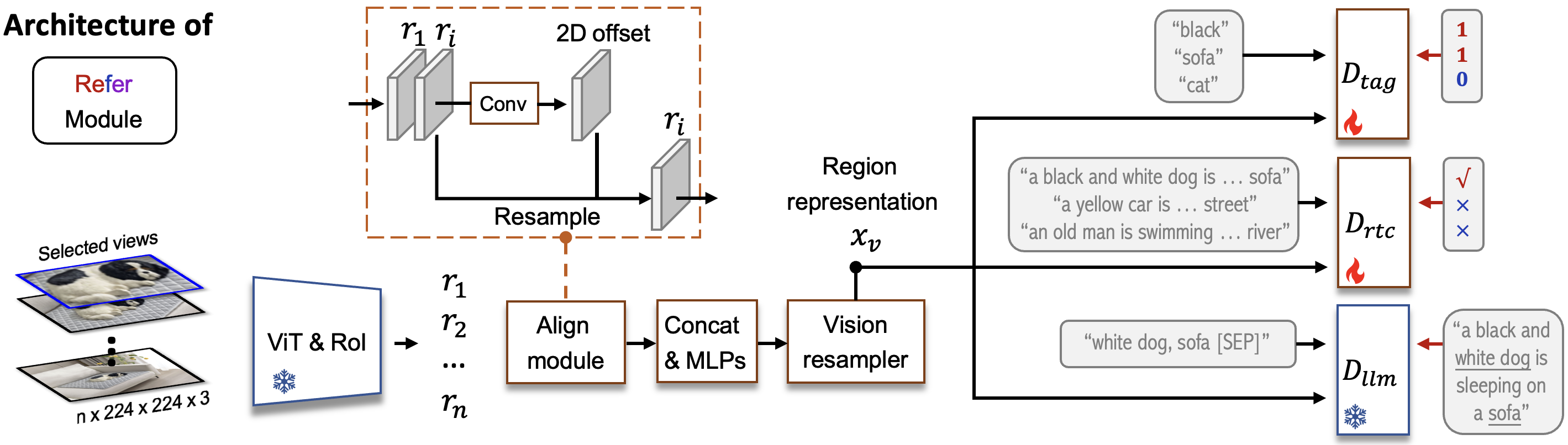}
 	\caption{Architecture of the proposed refer module. It comprises a stochastic multi-view embedding module and multimodal decoders ($D_{*}$). $n$ nested views are encoded as a region representation $x_v$ by the stochastic multi-view embedding module (left). The region representation $x_v$ is decoded by multimodal decoders, and then aligned to language descriptions of multimodal tasks (right).}
    \label{fig:3}
\end{figure*}

\section{Methodology}
\label{sec:method}

{The ``dynamic" capability is achieved through a stochastic vision-language alignment procedure during training and a selectively multimodal referring procedure during reference, Fig.~\ref{fig:2}}.
{In the training procedure,} we construct a set of nested image views that contain the referred region and randomly sample views to simulate an image with stochastic resolution.
A stochastic multi-view embedding procedure is then carried out to encode the image of stochastic resolution to a region representation, which is aligned to language descriptions of multimodal tasks.
{In the inference procedure,} the set of nested image views is constructed once again, and proper views are selected based on task prior and image prior, thereby improving the representational adaptability of region-level multimodal models.

\subsection{Training Dynamic Resolution: \\ Stochastic Vision-Language Alignment}
%
\subsubsection{Nested View Construction}
%
Vision foundation models, $e.g.$, CLIP and EVA-CLIP~\cite{openclip, fang2023eva}, are becoming more powerful, but remain handling fixed-resolution images.
To exploit their potential for encoding visual inputs of dynamic resolution, we seek a simple alternative by transforming the original image into multiple nested views that cover the referred regions, Fig.~\ref{fig:2}(left).
These nested views share the same resolution and can be combined to simulate an image with dynamic resolution, highlighting the referred region while depressing the irrelevant areas.

Specifically, the original image $x$ is cropped and resized into multiple candidate views. 
The cropped regions are calculated by $b_r + t * (b_x - b_r), t\in {\mathbb{R}} [0, 1]$. 
$b_r$, $b_x$, and $t$ respectively denote the bounding box of the referred region, the size of the whole image, and the interpolation coefficient. 
During training, $n$ views are stochastically sampled from the candidates to simulate images generated by foveation and saccadic eye movements. 
The $n$ views correspond to interpolation coefficients $\textbf{t}$, where $\textbf{t}=[t_1,t_2,\cdots,t_n]$. 
We keep the view containing only the referred region ($t_1=0$) being sampled, $i.e.$, the image with blue border in Fig.~\ref{fig:2}, which best preserves details and is experimentally validated crucial for all multimodal tasks.

\begin{figure*}[tbp]
	\includegraphics[width=1\linewidth]{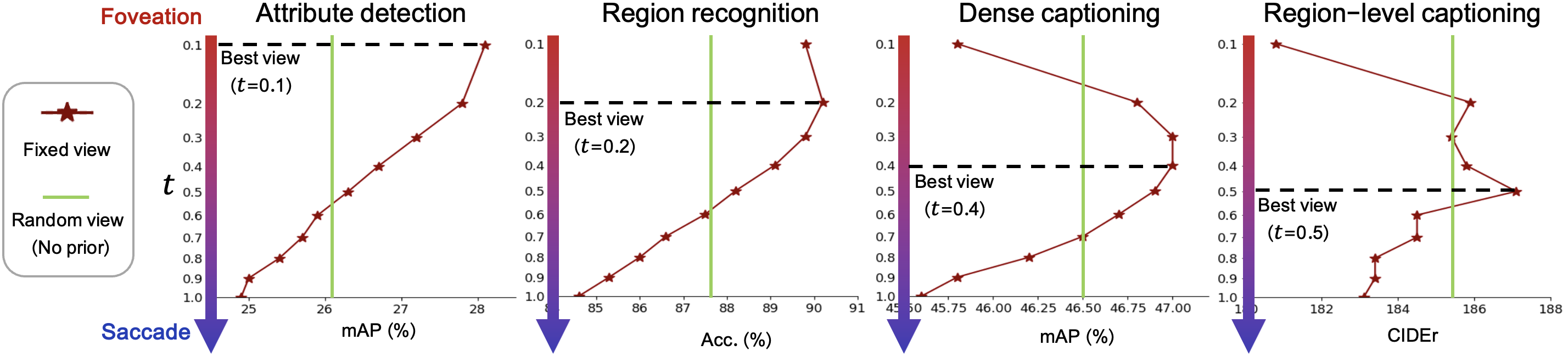}
 	\caption{Performance of a double-view ($n=2$) \Ours model on region-level multimodal tasks ($e.g.$, open-vocabulary attribute detection on OVAD~\cite{ovad}, region recognition on COCO~\cite{coco}, dense captioning on VG-COCO~\cite{Shao2022Region}, and region-level captioning on VG~\cite{krishna2017visual}) under interpolation coefficients $\textbf{t}$, $\textbf{t} = [t_1, t_2]\in {\mathbb{R}}^2[0, 1]$. The first view is a fixed one ($t_1=0$) and the second is randomly selected or fixed.}
    \label{fig:view}
\end{figure*}

\subsubsection{Stochastic Multi-view Embedding}
\label{sec:stochastic-multi-view-embedding}
The sampled $n$ views, $i.e.$, image with stochastic resolution, are jointly encoded by a frozen ViT into spatial features, which are further processed by an RoI-Align module~\cite{he2017mask} to obtain region embeddings, $i.e.$, $\{r_i\}_{i=1,2,\cdots, n}$, Fig.~\ref{fig:3}(left).
Due to biases introduced by cropping, resizing, and RoI-Align, the region embeddings are not well spatially aligned. 
Inspired by dynamic convolution operations~\cite{dcnv4,huang2021fapn}, we propose an align module (Fig.~\ref{fig:3} upper) to reduce the bias by aligning $\{r_i\}_{i=2,3,\cdots, n}$ to $r_1$, where $r_1$ is the region embedding that corresponds to the view containing only the referred region. 
Each region embedding $r_i$ is first concatenated with $r_1$, followed by a convolution layer to compute a 2D offset map.
The spatial feature of $r_i$ is then resampled according to the 2D offset. 
Finally, the aligned region embeddings are concatenated across the channel dimension and fused by a multi-layer perceptron (MLP) layer.
The outputs are further compressed by a vision resampler, $i.e.$, the Q-former~\cite{Li2023BLIP2}, so that we extract a region representation ($x_v$ in Fig.~\ref{fig:3}) for the referred region $b_r$ of the image $x$. 

\subsubsection{Vision-Language Alignment}
\label{vision-language-alignment}
The region representation $x_v$, calculated through the stochastic multi-view embedding process, is decoded by three decoders\footnote{Please refer to  Appendix~A for more details about the decoders.}. $D_{*}$ is shown in Fig.~\ref{fig:3}(right), which are respectively supervised by three multimodal tasks:

\textbf{i) Image Region Tagging.} 
Inspired by the off-the-shelf image tagging methods~\cite{query2label, tag2text, ram}, we apply a query-based lightweight recognition decoder~\cite{query2label} for region tagging.
The decoder $D_{tag}$ is shown in Fig.~\ref{fig:3}(right).  
This tagging procedure is fulfilled through calculating the confidence of predefined tags by using tags as query and $x_v$ as key and value, respectively. 
Following the control captioning method~\cite{zhao2024controlcap}, we parse the ground-truth tags from the captions to supervise the recognition decoder. 
To handle the problem of missing labels of regions, the asymmetric loss~\cite{ridnik2021asymmetric}, which is robust to imprecise supervision, is used for model optimization. 

\textbf{ii) Region-text Contrastive Learning.} 
Similar to the decoder for region tagging, the decoder $D_{rtc}$ is defined as a query-based recognition decoder~\cite{query2label}, which calculates the similarity scores between captions and region features by using the former as the query and the latter as the key and value. 
This is actually a contrastive learning procedure, where the similarity scores are optimized through the pairwise Sigmoid loss for Language-Image Pre-training~\cite{siglip}. 
Unlike standard contrastive learning with softmax normalization, the sigmoid loss operates solely on image-text pairs and does not require a global view of the pairwise similarities for normalization.

\textbf{iii) Language Modeling.} As shown in Fig.~\ref{fig:3}(right), a language modeling decoder $D_{llm}$ is used to convert region representation $x_v$ to language descriptions.
Following the typical design of LLMs~\cite{Li2023BLIP2, Liu2023LLaVa}, a learnable linear projector is used to map $x_v$ to the language space.
Together with the mapped $x_v$, random control embeddings\footnote{More details of the control embeddings are provided in Appendix~B.} ~\cite{zhao2024controlcap} built upon word pieces parsed from the ground-truth captions are fed to a frozen LLM for text generation. 
The language outputs are supervised by the ground-truth captions with a cross-entropy loss~\cite{zhao2024controlcap, Li2023BLIP2, Liu2023LLaVa}.

\subsection{Inference Dynamic Resolution: \\ Selectively Multimodal Referring}
\label{sec:dynamic-multimodal-referring}

During inference, the trained \Ours model performs multimodal referring on images with dynamic resolutions. 
By adjusting the interpolation coefficients $\textbf{t}$ ($\textbf{t}=[t_1,t_2,\cdots,t_n]$) for the sampled $n$ views, we obtain region representations with dynamic resolution characteristics.
This is consistent with the training procedure. 
The key challenge in multimodal referring is how to adjust the interpolation coefficients 
$\textbf{t}$ of the views to select the best view. To this end, we propose two solutions for the two distinct cases, Fig.~\ref{fig:2}(lower).

\textbf{i) Inference with task prior:} When task prior is known in advance, views are selected based on the specific attributes and characteristics of the task. To investigate the characteristics of existing region-level multimodal tasks, we train a double-view ($n=2$) \Ours model and evaluate it on four tasks. From the curves in Fig.~\ref{fig:view}, we can conclude that better results are achieved for attribute detection under contextless views ($t_2=0.1$), which refer to image views tightly bound to the referred region. This is understandable, as such tasks typically require detailed region-specific information.
For region-level captioning and dense captioning, context-rich views ($t_2=0.4$ or $t_2=0.5$) provide better results, as these tasks rely on a more comprehensive context for a fuller understanding of the referred region.
It is worth noting that views with excessive context ($t_2>0.5$) degrade performance across all tasks, as they introduce too much irrelevant information from outside the region of interest.
Thus, by understanding the characteristics of region-level multimodal tasks, one can choose the appropriate views that effectively encode the necessary region representation for each task.

\begin{figure}[tbp]
    \centering
	\includegraphics[width=1\linewidth]{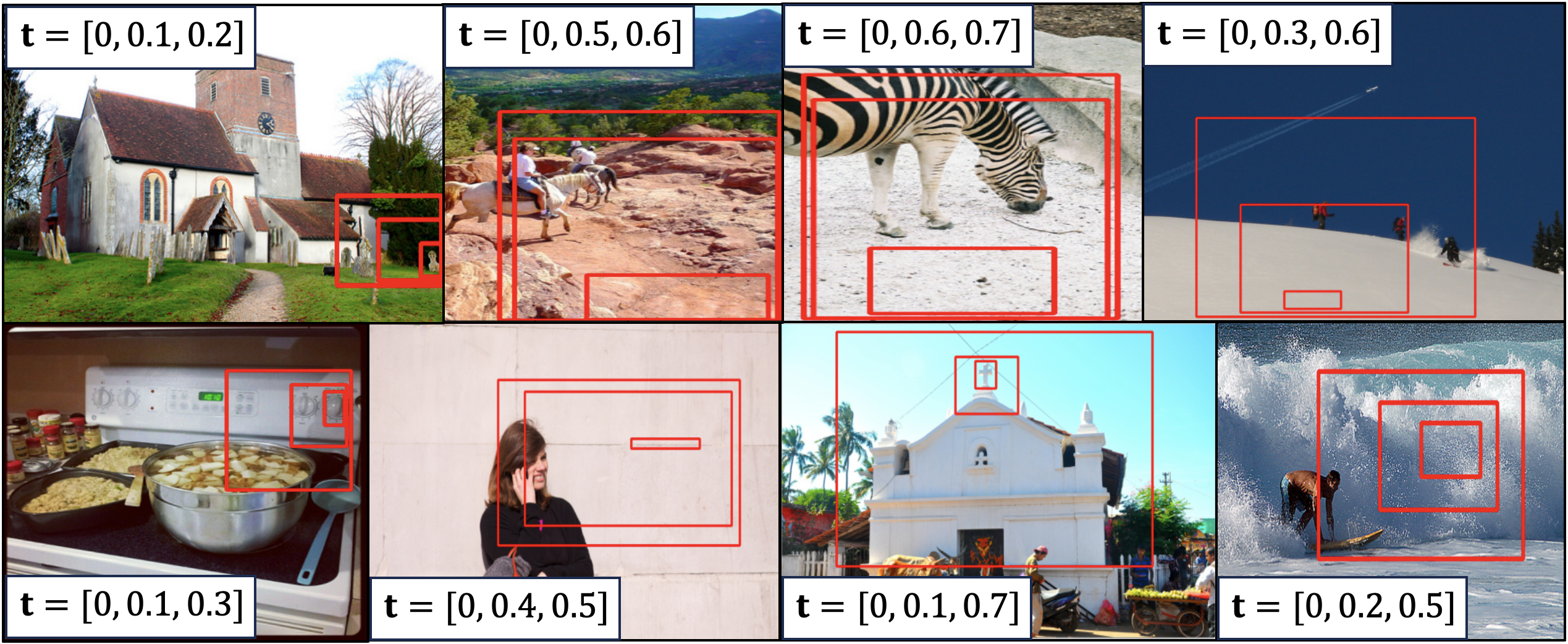}
    \caption{Visualization of selected views using image prior.}
    \label{fig:demo_image_prior}
\end{figure}
\vspace{-0.4mm}

\begin{table}[t]
    \footnotesize
    \centering
    \tabcolsep=0.07cm
    \caption{Region-level captioning performance of DynRefer and the state-of-the-art methods on the RefCOCOg and VG datasets.}
    \label{tab:performance_reg}
    \begin{tabular}{rc|c|c|c|c}
\toprule
\multirow{2}{*}{Method} & \multirow{2}{*}{\makecell{Model \\ size}} & \multicolumn{2}{c|}{RefCOCOg} & \multicolumn{2}{c}{VG}\tabularnewline
\cline{3-6} 
 &  & METEOR & CIDEr & METEOR & CIDEr\tabularnewline
\midrule
SLR+Rerank$_{\text{CVPR'17}}$~\cite{yu2017joint} & <1B & 15.9 & 66.2 & -  & -\tabularnewline
GPT4RoI$_{\text{ARXIV'23}}$~\cite{zhang2023gpt4roi} & 7.4B & - & - & 17.4 & 145.2\tabularnewline
GRiT$_{\text{ECCV'24}}$~\cite{Wu2022GRIT} & <1B & 15.2 & 71.6 & 17.1 & 142.0\tabularnewline
Groma$_{\text{ECCV'24}}$~\cite{ma2025groma} & 7.4B & 16.8 & 107.3 & 19.0 & 158.4\tabularnewline
ControlCap$_{\text{ECCV'24}}$~\cite{zhao2024controlcap} & 4.2B & 17.0 & 111.4 & 20.4 & 181.9\tabularnewline
Kosmos-2$_{\text{ICLR'24}}$~\cite{Peng2023Kosmos2} & 1.6B & 14.1 & 62.3 & -  & -\tabularnewline
RegionGPT$_{\text{CVPR'24}}$~\cite{glamm} & 7.4B & 16.9 & 109.9 & 17.0 & 145.6\tabularnewline
GLaMM$_{\text{CVPR'24}}$~\cite{glamm} & 7.4B & 16.2 & 106.0 & 19.7 & 180.5\tabularnewline
Alpha-CLIP$_{\text{CVPR'24}}$~\cite{sun2023alpha} & 7.4B & 16.7 & 109.2 & 18.9 & 160.3\tabularnewline
Osprey$_{\text{CVPR'24}}$~\cite{osprey} & 7.3B & 16.6 & 108.3 & - & -\tabularnewline
\midrule
\rowcolor{gray0} \Ours (Ours) & 4.2B & \textbf{18.1} & \textbf{115.7} & \textbf{21.2} & \textbf{190.9}\tabularnewline
\bottomrule
\end{tabular}
\end{table}
\vspace{-0.4mm}

\begin{table}[t]
    \footnotesize
    \centering
    \tabcolsep=0.12cm
    \caption{Open vocabulary attribute detection performance of DynRefer and the state-of-the-art methods on the OVAD dataset with the box-oracle setup (OVAD-Box).}
    \begin{tabular}{rc|c|c|c|c}
\toprule
\multirow{2}{*}{Method} & \multirow{2}{*}{Backbone} & \multicolumn{4}{c}{OVAD-Box}\tabularnewline
\cline{3-6}
& & All & Head & Medium & Tail\tabularnewline
\midrule
Chance~\cite{ovad} & - & 8.6 & 36.0 & 7.3 & 0.6\tabularnewline
CLIP$_{\text{ICML'21}}$~\cite{Radford2021CLIP} & ResNet50 & 15.8 & 42.5 & 17.5 & 4.2\tabularnewline
CLIP$_{\text{ICML'21}}$~\cite{Radford2021CLIP} & ViT-B16 & 16.6 & 43.9 & 18.6 & 4.4\tabularnewline
Open CLIP$_{\text{ICML'21}}$~\cite{openclip} & ResNet50 & 11.8 & 41.0 & 11.7 & 1.4\tabularnewline
Open CLIP$_{\text{ICML'21}}$~\cite{openclip} & ViT-B16 & 16.0 & 45.4 & 17.4 & 3.8\tabularnewline
Open CLIP$_{\text{ICML'21}}$~\cite{openclip} & ViT-B32 & 17.0 & 44.3 & 18.4 & 5.5\tabularnewline
ALBEF$_{\text{NeurIPS'21}}$~\cite{albef} & ViT-B16 & 21.0 & 44.2 & 23.9 & 9.4\tabularnewline
X-VLM$_{\text{ICML'22}}$~\cite{xvlm} & Swin-B & 28.1 & 49.7 & 34.2 & 12.9\tabularnewline
OVAD-Baseline$_{\text{CVPR'23}}$~\cite{ovad} & ViT-B32 & 21.4 & 48.0 & 26.9 & 5.2\tabularnewline
BLIP2$_{\text{ICML'23}}$~\cite{Li2023BLIP2} & EVA & 25.5 & 49.8 & 30.5 & 10.8\tabularnewline
\midrule
\rowcolor{gray0} \Ours (Ours) & ViT-L & 28.2 & 50.9 & 34.5 & 12.5\tabularnewline
\rowcolor{gray0} \Ours (Ours) & EVA & \textbf{29.2} & 49.9 & \textbf{35.7} & \textbf{14.0}\tabularnewline
\bottomrule
\end{tabular}
    \label{tab:performance_ovad}
\end{table}

\vspace{-0.4mm}

\textbf{ii) Inference with image prior:} When task prior is unknown, views are selected based on image priors by maximizing the total information provided by the combined views using a greedy search algorithm.
Specifically, we first construct a set of candidate views with different interpolation coefficients $t$, $t\in\{0.1,0.2,\cdots,1\}$. 
Among these candidates, the first view $x(t_1)$, which contains only the referred region, is always included.
In what follows, the remaining $n-1$ views are selected using a greedy search algorithm. The objective function for the search is formulated as:
\begin{align}
& \mathop{\text{argmax}}\limits_{t_i} \frac{\sum (\text{pHASH}(x(t_1)) \oplus \text{pHASH}(x(t_i)))}{t_i},
\end{align}
where $x(t_i)$ represents the $i$-th view and $t_i$ denotes its interpolation coefficient.
The term ``$\sum (\text{pHASH}(x(t_1)) \oplus \text{pHASH}(x(t_i)))$'' quantifies the incremental information introduced by the $i$-th view compared to the first view. 
The perceptual image hash function, ``$\text{pHASH}(\cdot)$'' encodes the views into hash codes in the frequency domain, and the XOR operation ``$\oplus$'' is applied to compare the hash codes of the $x(t_1)$ and $x(t_i)$, capturing the difference in information.
The factor ``$\frac{1}{t_i}$'' serves to downweight context-rich views ($t_i>0.5$), reducing the risk of introducing redundant information, as observed in Fig.~\ref{fig:view}. This ensures that the search prioritizes views with balanced and informative content~\footnote{Please refer to Appendix~C for the illustration of pHASH operation.}. The visualizations of selected views based on the image prior are shown in Fig.~\ref{fig:demo_image_prior}. When the referred region contains minimal information ($e.g.$, a white wall or the ground), the proposed strategy selects more informative views to complement the region.

\begin{table}[t]
    \footnotesize
    \centering
    \tabcolsep=0.09cm
    \caption{Dense captioning performance of DynRefer and the state-of-the-art methods on the VG and VG-COCO datasets. When requiring localization, we use a pre-trained GRiT~\cite{Wu2022GRIT} model to provide bounding boxes.}
    \label{tab:performance_dense_caption}
    \begin{tabular}{rc|c|c|c}
\toprule
\multirow{2}{*}{Method} & \multirow{2}{*}{\makecell{GT \\ localization}} & \multicolumn{3}{c}{mAP(\%)}\tabularnewline
\cline{3-5} 
 &  & VG V1.0 & VG V1.2 & VG-COCO\tabularnewline
\midrule
FCLN$_{\text{CVPR'16}}$~\cite{Johnson2016DenseCap} & \ding{55} & 5.4 & 5.2 & -\tabularnewline
JIVC$_{\text{CVPR'17}}$~\cite{Yang2017Dense} & \ding{55} & 9.3 & 10.0 & -\tabularnewline
ImgG$_{\text{AAAI'19}}$~\cite{Li2019Learning} & \ding{55} & 9.3 & 9.7 & -\tabularnewline
COCD$_{\text{AAAI'19}}$~\cite{Li2019Learning} & \ding{55} & 9.4 & 9.8 & 7.9\tabularnewline
COCG$_{\text{AAAI'19}}$~\cite{Li2019Learning} & \ding{55} & 9.8 & 10.4 & 8.9\tabularnewline
CAG-Net$_{\text{CVPR'19}}$~\cite{Yin2019Context} & \ding{55} & 10.5 & - & -\tabularnewline
TDC$_{\text{TNNLS'22}}$~\cite{Shao2022Region} & \ding{55} & 11.5 & 11.9 & 11.9\tabularnewline
CapDet$_{\text{CVPR'23}}$~\cite{Long2023CapDet} & \ding{55} & - & 15.4 & 14.0\tabularnewline
DCMSTRD$_{\text{TMM'24}}$~\cite{10444947} & \ding{55} & 13.6 & 13.4 & 16.1\tabularnewline
GRiT$_{\text{ECCV'24}}$~\cite{Wu2022GRIT} & \ding{55} & 15.5 & 16.4 & -\tabularnewline
ControlCap$_{\text{ECCV'24}}$~\cite{zhao2024controlcap} & \ding{55} & 18.2 & 18.5 & 18.4\tabularnewline
PixelLLM$_{\text{CVPR'24}}$~\cite{ren2024pixellm} & \ding{55} & 17.0 & - & -\tabularnewline
DetCLIPv3$_{\text{CVPR'24}}$~\cite{detclipv3} & \ding{55} & - & \textbf{19.7} & 18.9\tabularnewline
\rowcolor{gray0} \Ours (Ours) & \ding{55} & \textbf{19.1} & 19.5 & \textbf{19.4}\tabularnewline
\midrule
FCLN$_{\text{CVPR'16}}$~\cite{Johnson2016DenseCap} & \ding{51} & 27.0 & - & -\tabularnewline
JIVC$_{\text{CVPR'17}}$~\cite{Yang2017Dense} & \ding{51} & 33.6 & - & -\tabularnewline
CAG-Net$_{\text{CVPR'19}}$~\cite{Yin2019Context} & \ding{51} & 36.3 & - & -\tabularnewline
BLIP2$_{\text{ICML'23}}$~\cite{Li2023BLIP2} & \ding{51} & 37.7 & 37.9 & 36.9\tabularnewline
GRiT$_{\text{ECCV'24}}$~\cite{Wu2022GRIT} & \ding{51} & 40.0 & 40.3 & -\tabularnewline
ControlCap$_{\text{ECCV'24}}$~\cite{zhao2024controlcap} & \ding{51} & 42.4 & 42.8 & 43.2\tabularnewline
\rowcolor{gray0} \Ours (Ours) & \ding{51} & \textbf{47.2} & \textbf{47.4} & \textbf{47.6}\tabularnewline
\bottomrule
\end{tabular}  
\end{table}


\begin{table}[t]
    \footnotesize
    \centering
    \tabcolsep=0.09cm
    \caption{Open vocabulary region recognition performance of DynRefer and state-of-the-art methods on the COCO-2017 val set. Following RegionGPT~\cite{guo2024regiongpt} and RegionCLIP~\cite{regionclip}, we report the results of object classification given ground-truth boxes.}
    \begin{tabular}{rcc|cc}
\toprule
Method & Backbone & LLM & mAP & Acc. (\%)\tabularnewline
\midrule
CLIP$_{\text{ICML'21}}$~\cite{Radford2021CLIP} & ViT-L & - & 58.9 & -\tabularnewline
RegionCLIP$_{\text{CVPR'22}}$~\cite{regionclip} & R50 & - & 58.3 & -\tabularnewline 
LLaVA$_{\text{NeurIPS'23}}$~\cite{Liu2023LLaVa} & ViT-L & Vicuna-7B & - & 40.0\tabularnewline
Shikra$_{\text{ARXIV'23}}$~\cite{Chen2023Shikra} & ViT-L & Vicuna-7B & - & 53.9\tabularnewline
GPT4RoI$_{\text{ARXIV'23}}$~\cite{zhang2023gpt4roi} & ViT-L & LLaVA-7B & - & 64.0\tabularnewline 
PVIT$_{\text{ARXIV'23}}$~\cite{pvit} & ViT-L+R50 & LLaVA-7B & - & 64.5\tabularnewline 
ASM$_{\text{ICLR'24}}$~\cite{wang2023allseeing} & ViT-L & Hasky-7B & 69.3 & -\tabularnewline
RegionGPT$_{\text{CVPR'24}}$~\cite{guo2024regiongpt} & ViT-L & Vicuna-7B & 70.0 & 80.6\tabularnewline
\midrule
\rowcolor{gray0} \Ours (Ours) & ViT-L & FlanT5$_{\text{XL}}$-3B & 85.0 & 89.4\tabularnewline
\rowcolor{gray0} \Ours (Ours) & EVA & FlanT5$_{\text{XL}}$-3B & \textbf{89.2} & \textbf{91.8} \tabularnewline
\bottomrule
\end{tabular}
    \label{tab:performance_coco}
\end{table}

\section{Experiment}
\label{sec:experiment}

\begin{figure*}[tbp]
    \centering
	\includegraphics[width=1\linewidth]{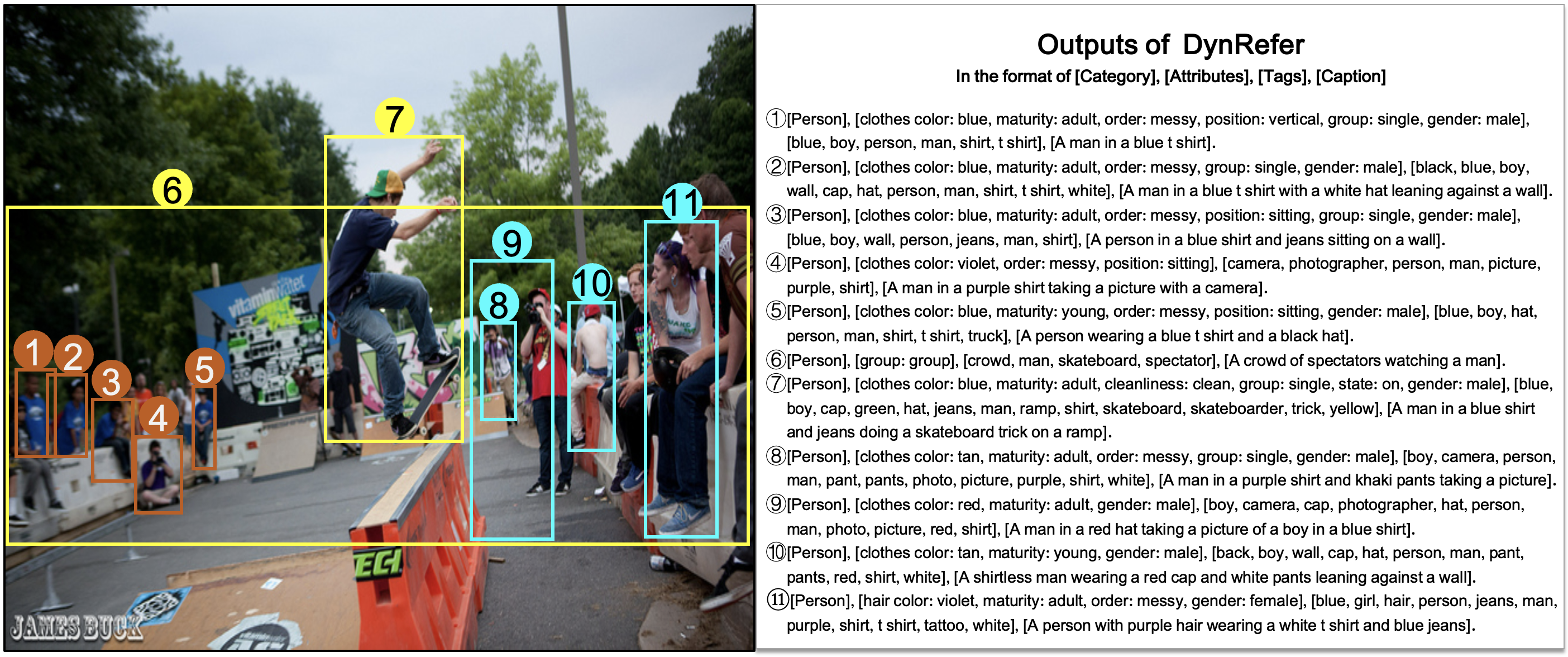}
    \caption{Illustration of \Ours's multi-task capability. It can generate captions, tags, attributes, categories for any referred regions.}
    \label{fig:demo}
\end{figure*}

\begin{table}[t]
    \footnotesize
    \centering
    \tabcolsep=0.09cm
    \caption{Referring reasoning performance of Finetuned DynRefer and the state-of-the-art methods on the Ferret-Bench~\cite{you2023ferret}.}
    \label{tab:performance_vqa}
    \begin{tabular}{rc|c}
\toprule
Method & Model size & Referring Reasoning\tabularnewline
\midrule
Shikra-7B$_{\text{ARXIV'23}}$~\cite{Chen2023Shikra} & 7.4B & 41.6\tabularnewline
Kosmos-2$_{\text{ICLR'24}}$~\cite{Peng2023Kosmos2} & 1.6B & 33.7\tabularnewline
Ferret-7B$_{\text{CVPR'24}}$~\cite{you2023ferret} & 7.4B & 67.3\tabularnewline
Osprey$_{\text{CVPR'24}}$~\cite{osprey} & 7.3B & 67.8\tabularnewline
\midrule
\rowcolor{gray0} \Ours (Ours) & 4.2B & \textbf{68.9}\tabularnewline
\bottomrule
\end{tabular}  
\end{table}

\Ours is implemented upon the LAVIS~\cite{li2022lavis} framework, where vision transformer, vision resampler and large language model are respectively initialized by EVA~\cite{fang2023eva}, Q-former~\cite{Li2023BLIP2} and FlanT5$_{\text{XL}}$~\cite{Chung2022Flan5} by default unless otherwise specified.
All the sampled views are resized to $224\times224$ resolution.
All models can be trained less than 20 hours using 8 NVIDIA A800 GPUs. For performance comparison, we train a triple-view ($n=3$) \Ours model and inference with image prior on VG V1.2~\cite{krishna2017visual} ($i.e.$, results in Tab.~\ref{tab:performance_reg}~\ref{tab:performance_dense_caption}~\ref{tab:performance_ovad}~\ref{tab:performance_coco}~\ref{tab:performance_fps}).
%
For ablation studies, \Ours is trained on VG-COCO~\cite{Shao2022Region} ($i.e.$, results in Tab.~\ref{tab:abltion_smvr}). For results on RefCOCOg~\cite{yu2016modeling}, \Ours is finetuned on its training set.
For referring reasoning, \Ours is finetuned on the combination of LLaVA and Osprey instruction tuning datasets.
Please refer to Appendix~D for more details about model/dataset/evaluation settings.

\subsection{Performance}

\noindent\textbf{Region-level Captioning.} In Tab.~\ref{tab:performance_reg} and Tab.~\ref{tab:performance_dense_caption}, 
\Ours is compared with the state-of-the-art (SOTA) methods.
\Ours respectively achieves 18.1 and 21.2 METEOR scores, 115.7 and 190.9 CIDEr scores on RefCOCOg and VG, outperforming the SOTA methods with a much smaller model size (4.2B $vs.$ 7B).
For dense captioning, \Ours achieves comparable performance with DetCLIPv3~\cite{detclipv3}.
When the ground-truth localization is given, \Ours respectively achieves 47.2\%, 47.4\% and 47.6\% mAPs on VG V1.0, V1.2, and VG-COCO, outperforming GRiT~\cite{Wu2022GRIT} by large margins.  

\noindent\textbf{Open-Vocabulary Attribute Detection.} The performance is shown in Tab.~\ref{tab:performance_ovad}. \Ours achieves 29.2\% mAP on OVAD, outperforming the SOTA methods. On Medium and Tail attributes, \Ours achieves the highest mAP, which demonstrates the generalizability of the proposed approach.

\noindent\textbf{Open-Vocabulary Region Recognition.} The performance is shown in Tab.~\ref{tab:performance_coco}. \Ours outperforms the SOTA methods by large margins (up to 8.8\% Acc. and 15\% mAP) with a smaller language model. 

\begin{figure}[tbp]
    \centering
	\includegraphics[width=1\linewidth]{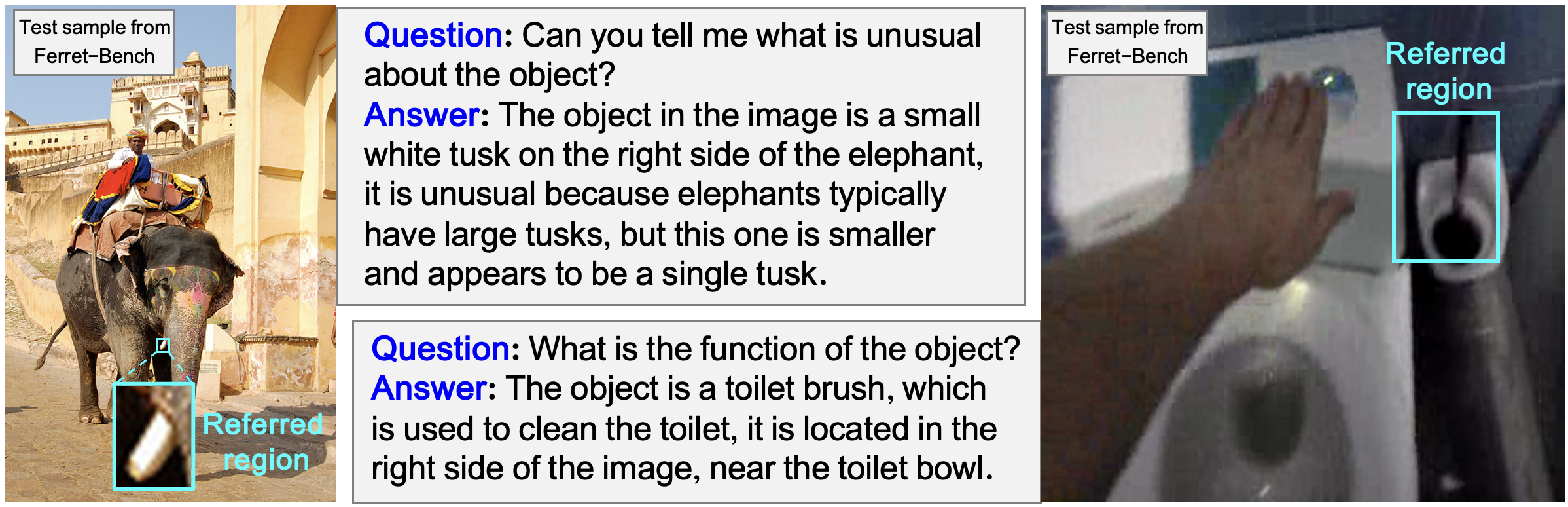}
    \caption{Illustration of finetuned \Ours's referring reasoning capability.}
    \label{fig:vqa_demo}
\end{figure}

\begin{table*}[t]
    \small
    \centering
    \caption{Comparison of FLOPs of vision encoder (Vis. FLOPs), inference speed (FPS, Frame Per Second), and region-level captioning performance on RefCOCOg between GLaMM~\cite{glamm}, Osprey~\cite{osprey} and the proposed approach. FPS is tested on a single A100 GPU. We omit the FLOPs of the language model as it varies with the length of generated sequence. Besides, \Ours is equipped with FlanT5$_{\text{XL}}$-3B, which is smaller and more efficient than Vicuna-7B.}
    \begin{tabular}{rcc|c|c|c|c}
\toprule 
Method & Backbone & LLM & Vis. FLOPs & FPS & METEOR & CIDEr\tabularnewline
\midrule
GLaMM$_{\text{CVPR'24}}$~\cite{glamm} & ViT-L (Res. 336) & Vicuna-7B & 257G & 0.71 & 16.2 & 106.0\tabularnewline
Osprey$_{\text{CVPR'24}}$~\cite{osprey} & ConvNext-L (Res. 512) & Vicuna-7B & 200G &2.58  & 16.6 & 108.3\tabularnewline
\midrule
\rowcolor{gray0} \Ours (Ours) & ViT-L (Res. 224 + Stochastic 3-view) & Vicuna-7B & 206G & 2.43 & 17.4 & 110.7  \tabularnewline
\rowcolor{gray0} \Ours (Ours) & ViT-L (Res. 224 + Stochastic 3-view) & FlanT5$_{\text{XL}}$-3B & 206G &4.81  & 17.3 & 109.7\tabularnewline
\rowcolor{gray0} \Ours (Ours) & EVA (Res. 224 + Stochastic 2-view) & FlanT5$_{\text{XL}}$-3B & 530G & 2.55 & 17.9 & 114.7\tabularnewline
\bottomrule
\end{tabular}

    \label{tab:performance_fps}
\end{table*}
\vspace{-0.4mm}

\begin{table*}[t]
    \small
    \centering
    \tabcolsep=0.15cm
    \caption{Ablation studies of \Ours on region-level multimodal benchmarks. Line 1: Training with cropped images~\cite{Li2023BLIP2}. Line 2: Training with images with RoI-Align~\cite{zhang2023gpt4roi,glamm}. Lines 3-4: Training with higher resolution images. Line 5: Training with fixed 2-view~\cite{zhao2024controlcap}. Lines 6-11: Training with Stochastic $n$-view (Ours). 
    Lines 12-18: The effect of removing some core modules or designs in \Ours.
    For model inference, ``No prior'', ``Task prior'', ``Image prior'' respectively denotes inference with randomly selected views, selection strategies based on task prior and image prior proposed in Sec.~\ref{sec:dynamic-multimodal-referring}. More details of the ablation studies are provided in Appendix~E.}
    
    \label{tab:abltion_smvr}
    \begin{tabular}{c|cc|c|c|c|c|c|c}
\toprule
\multirow{2}{*}{} & \multirow{2}{*}{Training} & \multirow{2}{*}{Inference} & \multirow{2}{*}{Vis. FLOPs} & OVAD & COCO & VG-COCO & \multicolumn{2}{c}{RefCOCOg} \tabularnewline
\cline{5-9}
 &  & & & mAP (\%) & Acc. (\%) & mAP (\%) & METEOR & CIDEr \tabularnewline
\midrule
1 & Cropped image & - & 268G & 23.0 & 77.0 & 40.0 & 17.1 & 107.3\tabularnewline
2 & Image + RoIAlign & - & 268G & 19.7 & 74.3 & 39.1 & 17.2 & 110.0\tabularnewline
3 & Line 2 + Res.~224 $\rightarrow$ 336 & - & 618G & 21.7 & 80.1& 41.5 & 17.4 & 111.3 \tabularnewline
4 & Line 2 + Res.~224 $\rightarrow$ 448 & - &  1146G & 22.7 & 81.2 & 41.8 & 17.3 & 113.0 \tabularnewline
\midrule
5 & Fixed 2-view & - & 530G & 25.4 & 85.4 & 45.8 & 17.9 & 114.2 \tabularnewline
6 & Stochastic 2-view & No prior & 530G & 26.1 & 87.8 & 46.6 & 17.9 & 114.4 \tabularnewline
7 & Stochastic 2-view & Image prior & 530G & 27.5 & 89.3 & 46.8 & 17.9 & 114.7  \tabularnewline
8 & Stochastic 2-view & Task prior & 530G & 28.1 & 90.2 & 47.0 & 18.1 & 115.6\tabularnewline
9 & Stochastic 3-view & No prior & 792G & 27.3 & 88.9 & 47.3 & 18.2 & 117.7\tabularnewline
10 & Stochastic 3-view & Image prior & 792G  & 28.7 & 90.3 & \textbf{47.4} & \textbf{18.2} & \textbf{118.6}\tabularnewline
11 & Stochastic 4-view & Image prior & 1054G  & 27.2 & \textbf{90.7} & 47.1 & 17.8 & 114.1 \tabularnewline
\midrule
12 & Line 10 - $D_{llm}$ & Image prior & 792G  & 27.6 & 89.0 & - & - & - \tabularnewline
13 & Line 10 - $D_{rtc}$ & Image prior & 792G  & - & - & 47.0 & 18.1 & 114.2\tabularnewline
14 & Line 10 - $D_{tag}$ & Image prior & 792G   & 27.0 & 90.3 & 44.8 & 16.7 & 118.4\tabularnewline
15 & Line 10 - Pretrained Q-former & Image prior & 792G   & \textbf{28.8} & 90.3  & 46.7 & 17.9 & 113.1\tabularnewline
16 & Line 10 - ($t_1=0$) & Image prior & 792G & 23.0 & 74.0 & 44.4 & 17.2 & 110.0\tabularnewline
17 & Line 10 - Nesting views & No prior & 792G & 25.5 & 83.7  & 43.3 & 17.5 & 114.0 \tabularnewline
18 & Line 10 - Align module & No prior & 792G & 27.3 & 88.4 & 47.1 & 17.9 & 113.6\tabularnewline
\bottomrule
\end{tabular}
\end{table*}
\vspace{-0.4mm}

\noindent\textbf{Referring Reasoning.} The performance is shown in Tab.~\ref{tab:performance_vqa}. \Ours outperforms Osprey~\cite{osprey} by 1.1 (68.9 $vs.$ 67.8) with smaller model size (4.2B $vs.$ 7B). Some reasoning results of \Ours are shown in Fig.~\ref{fig:vqa_demo}.

\noindent\textbf{Computational Cost and Inference Speed.} As shown in Tab.~\ref{tab:performance_fps}, although \Ours processes more image views, the FLOPs of its vision encoder are comparable to GLaMM~\cite{glamm} and Osprey~\cite{osprey}, as each view has a low resolution. \Ours achieves better region-level captioning performance, with higher METEOR scores on RefCOCOg (17.9 $vs.$, 16.2) and faster inference speed compared to GLaMM~\cite{glamm} (2.55 $vs.$ 0.71). These results demonstrate that \Ours is efficient and suitable for real-world applications.

\subsection{Ablation Studies}
\label{sec:ablation}

\noindent\textbf{Stochastic Multi-view Embedding.} In Tab.~\ref{tab:abltion_smvr}, we compare the proposed stochastic multi-view embedding approach with other commonly used region representation methods~\cite{Li2023BLIP2, zhang2023gpt4roi, zhao2024controlcap}. 
In lines 1-2, the model is trained with resolution-fixed images. 
In lines 3-4, we increase the resolution of input images based on line 2 following common practice~\cite{osprey}, which brings higher FLOPs and limited performance gain. 
In line 5, the model is trained with visual input of fixed 2-view, which has acceptable FLOPs and large performance gain, demonstrating the efficiency of encoding images of dynamic resolution.
In lines 6-8, the views are stochastically sampled during training, resulting in performance gains across all tasks without extra computational cost, demonstrating the effectiveness of simulating the mechanism of foveation and saccade in human cognition.
In lines 9-10, we increase the number of sampled views to 3, which further improves performance at an acceptable cost in terms of FLOPs. However, in line 11, increasing the number of views to 4 leads to a performance drop. This is because with 4 views, there are $C_{10}^3$ possible combinations of views, which makes the manifold of region representations too complex and harder to optimize.

\noindent\textbf{Vision-Language Alignment.} The effectiveness of aligning region representations of images to language descriptions of multi-tasks is validated in lines 12-14, Tab.~\ref{tab:abltion_smvr}. Dropping any decoder results in performance degradation, highlighting the mutual improvements among tasks. Aligning without the pretrained Q-former slightly reduces the performance as shown in line 15. Unfixing the view ($t_1=0$) significantly harms performance as shown in line 16, demonstrating the importance of the view containing only the referred region, which best preserves details. In line 17, instead of using the strategy of nesting views in Sec.~\ref{sec:stochastic-multi-view-embedding}, we randomly select views that contains the referred region during training and inference, which deteriorates the performance. Finally, removing the align module in Fig.~\ref{fig:3} slightly reduces the performance on all tasks.

\noindent\textbf{Selectively Multimodal Referring.} The effectiveness of selecting views based on task prior and image prior is evaluated in lines 6-10 in Tab.~\ref{tab:abltion_smvr}. Compared to randomly selected views (No prior) during inference, selecting views based on task prior improve the performance across all tasks as shown in lines 6-8. When task prior is unavailable during inference, selecting views based on image prior is a useful alternative as shown in lines 7-10. 
\section{Conclusion}
\label{sec:conclusion}

We present \Ours, a resolution-adaptive approach to pursue high-accuracy region-level referring through mimicking the resolution adaptability of human visual cognition.
With stochastic vision-language alignment and selectively multimodal referring, \Ours predicts desired language descriptions for multimodal tasks, as well as customizing the resolution of referred image regions according to the task and image priors.
With its powerful adaptability, \Ours improves the performance of region-level multimodal tasks, with striking contrast to the state-of-the-art methods.
Furthermore, \Ours provides a fresh insight to unify region-level multimodal tasks.

{
    \small
    \bibliographystyle{ieeenat_fullname}
    \bibliography{main}
}

\newpage
\appendix

\begin{figure*}[ht]
	\centering\includegraphics[width=0.8\linewidth]{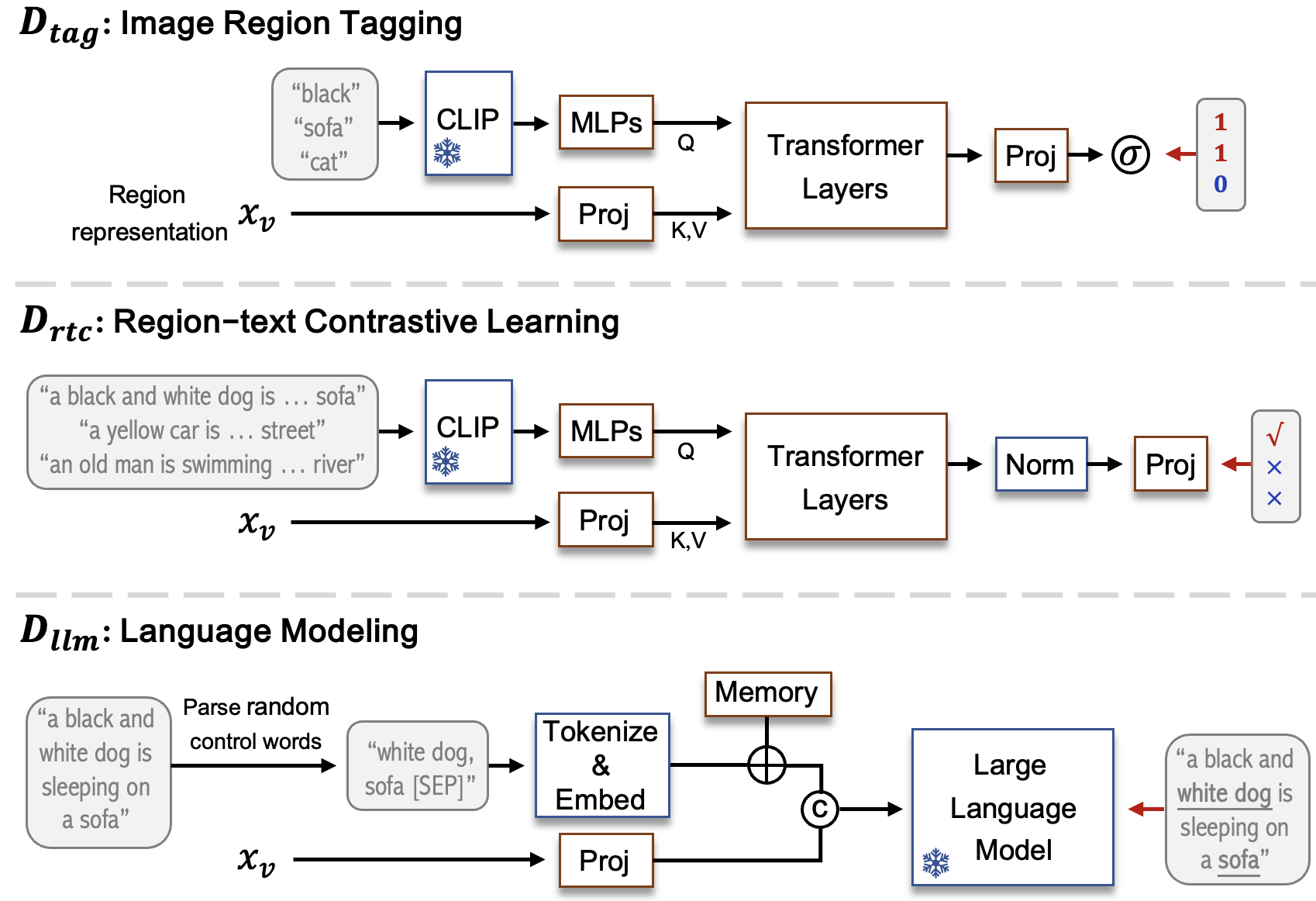}
     \caption{The detailed structure of multimodal decoders $D_{*}$ of \Ours. ``Proj'' is a linear projection layer. ``$\sigma$'' is the sigmoid activation function. ``Memory'' is a learnable embedding. The ``Transformer Layers'' denotes query-based decoders~\cite{query2label, ram} that contains only cross-attention layers and feed forward networks.}
    \label{fig:decoder}
\end{figure*}

\section{Structure of the Decoders in \Ours}
\label{app:decoders}

The structure of the decoders in \Ours is shown in Fig.~\ref{fig:decoder}.

\begin{figure}[ht]
	\centering\includegraphics[width=0.8\linewidth]{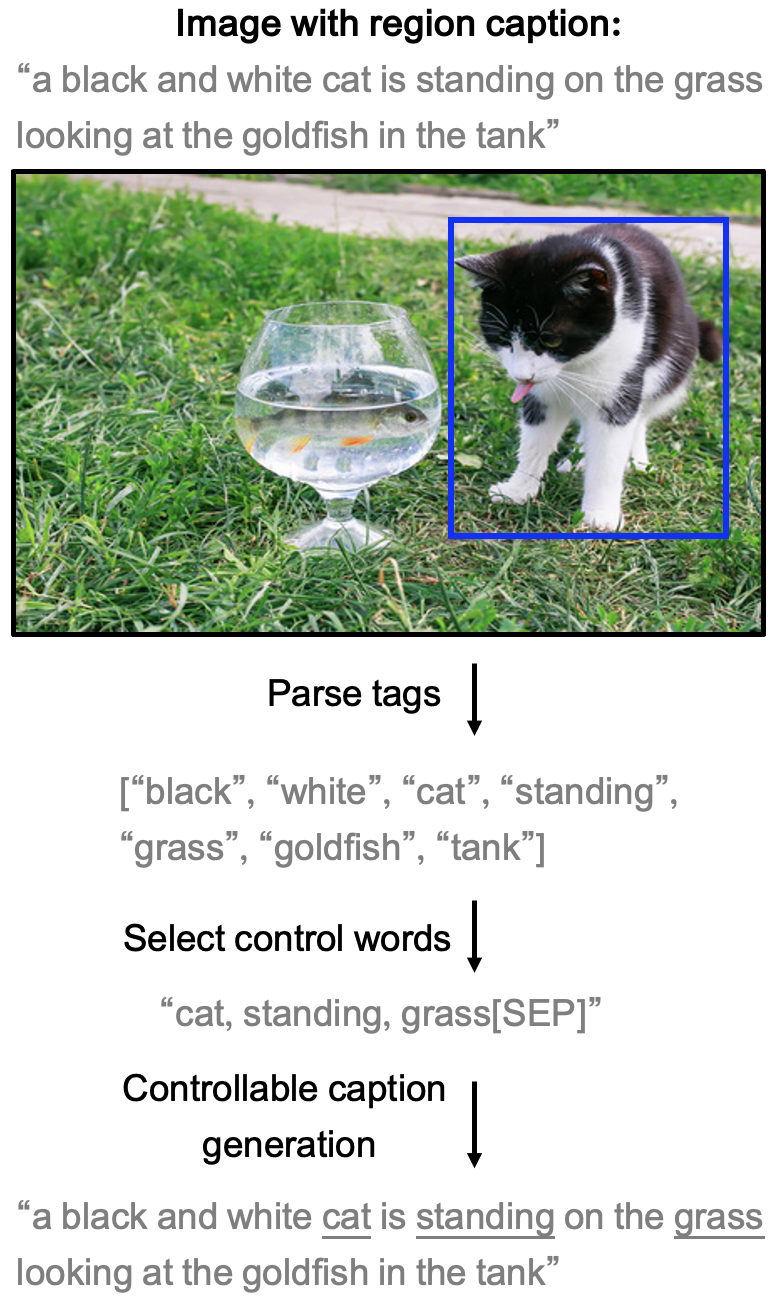}
     \caption{Illustration of the generation process of tags and control words used in \Ours.}
    \label{fig:supp_control_words}
\end{figure}

\textbf{i) Image Region Tagging.} As shown in Fig.~\ref{fig:decoder}($D_{tag}$), the region representation $x_v$ is first mapped to a low-dimension embedding with a linear projection layer. Meanwhile, predefined 4585 tags are encoded by a frozen CLIP~\cite{openclip} text encoder and multi-layer perceptrons. Then, a query-based decoder~\cite{query2label, ram} (``Transformer layers'' in Fig.~\ref{fig:decoder}) is used to calculate the confidences of the tags. The ground-truth tags are parsed from the caption of the referred region as shown in Fig.~\ref{fig:supp_control_words}. Finally, the confidences of the tags are optimized by asymmetric loss~\cite{ridnik2021asymmetric}, which is robust to imprecise supervision.

\textbf{ii) Region-text Contrastive Learning.}  As shown in Fig.~\ref{fig:decoder}($D_{rtc}$), it has a similar structure to $D_{tag}$. $D_{rtc}$ normalizes the outputs from the query-based decoder and projects them into similarity scores, which are optimized by the pairwise Sigmoid loss for Language-Image Pre-training~\cite{siglip}. 

\textbf{iii) Language Modeling.}  As shown in Fig.~\ref{fig:decoder}($D_{llm}$), following ControlCap~\cite{zhao2024controlcap}, random control words parsed from the ground-truth captions are combined to a sentence, $i.e.$, ``$\texttt{white dog, sofa[SEP]}$''. The sentence is encoded into the control embedding by the tokenizer and word embedding layer of the large language model. After that, a learnable memory unit is added to the control embedding. Finally, the control embedding and the projected region representation are concatenated and jointly sent into the large language model for text generation.

\section{Inference with Trained Decoders.} 
With trained decoders, the region representation $x_v$ can be decoded into region-level language descriptions, including tags, categories, attributes and captions. Their production are elaborated below:

i) \textit{tags}. The tags of the region are generated by $D_{tag}$. Following ~\cite{tag2text, ram}, we use a set of 4585 tags. During inference, we first query the decoder with the predefined tags to get the confidences. Then, the tags are filtered by a predefined tagging threshold.

ii) \textit{categories}. The category of the region is generated by $D_{rtc}$. During inference, we query the decoder with the template ``\texttt{a photo of a \{cls\}}'' and select the category with the highest score.

iii) \textit{attributes}. The attributes of the region are generated by $D_{rtc}$. During inference, we first query the decoder with attribute templates following OVAD~\cite{ovad}, $e.g.$, ``\texttt{the object has \{attr\}}''. Then, attributes with high scores are selected as the results.

iv) \textit{captions}. The caption of the region is generated by $D_{llm}$. During inference, we first use the tags of high confidence to form a control sentence, $i.e.$, ``\texttt{\{tag1\}, \{tag2\}, \{tag3\}, $\cdots$, [SEP]}''. Then, the control sentence and the region representation are used to control the language language model for caption generation.

\begin{table*}[t]
    \small
    \centering
    \tabcolsep=0.15cm
    \caption{Evaluation of the align module of \Ours on region-level multimodal benchmarks.}
    \begin{tabular}{c|cc|c|c|c|c|c|c}
\toprule
\multirow{2}{*}{} & \multirow{2}{*}{Align module} & \multirow{2}{*}{Inference} & \multirow{2}{*}{Vis. FLOPs} & OVAD & COCO & VG-COCO & \multicolumn{2}{c}{RefCOCOg} \tabularnewline
\cline{5-9}
 &  & & & mAP (\%) & Acc (\%) & mAP (\%) & CIDEr & METEOR \tabularnewline
\midrule
1 & \ding{55} & No prior & 790G & 27.3 & 88.4  & 47.1 & 17.9 & 113.6   \tabularnewline
2 & \ding{51} & No prior & 792G & 27.3 & 88.9 & 47.3 & 18.2 & 117.7 \tabularnewline
3 & \ding{55} & Image prior & 790G & 28.1 & \textbf{90.5}  & 46.6 & 17.7 & 113.5   \tabularnewline
4 & \ding{51} & Image prior & 792G & \textbf{28.7} & 90.3 & \textbf{47.4} & \textbf{18.2} & \textbf{118.6} \tabularnewline
\bottomrule
\end{tabular}
    \label{tab:abltion_align_module}
\end{table*}

\begin{table*}[t]
    \small
    \centering
    \tabcolsep=0.15cm
    \caption{Evaluation of the inference strategy of \Ours on region-level multimodal benchmarks.}
    \begin{tabular}{c|cc|c|c|c|c|c|c}
\toprule
\multirow{2}{*}{} & \multirow{2}{*}{Training} & \multirow{2}{*}{Inference} & \multirow{2}{*}{Vis. FLOPs} & OVAD & COCO & VG-COCO & \multicolumn{2}{c}{RefCOCOg} \tabularnewline
\cline{5-9}
 &  & & & mAP (\%) & Acc (\%) & mAP (\%) & CIDEr & METEOR \tabularnewline
\midrule
1 & Stochastic 2-view & No prior & 530G & 26.1 & 87.8 & 46.6 & 17.9 & 114.4 \tabularnewline
2 & Stochastic 2-view & Image prior & 530G & 27.5 & 89.3 & 46.8 & 17.9 & 114.7  \tabularnewline
3 & Stochastic 2-view & Task prior & 530G & 28.1 & 90.2 & 47.0 & 18.1 & 115.6\tabularnewline
4 & Stochastic 3-view & No prior & 792G & 27.3 & 88.9 & 47.3 & \textbf{18.2} & 117.7\tabularnewline
5 & Stochastic 3-view & Image prior & 792G  & 28.7 & 90.3 & \textbf{47.4} & \textbf{18.2} & \textbf{118.6}\tabularnewline
6 & Stochastic 3-view & Task prior & 792G  & \textbf{29.4} & \textbf{90.4} & \textbf{47.4} & \textbf{18.2} & 118.3
\tabularnewline
\bottomrule
\end{tabular}
    \label{tab:abltion_task_prior}
\end{table*}

\begin{table*}[t]
    \small
    \centering
    \tabcolsep=0.15cm
    \caption{Analysis of parameter composition of \Ours. Modules that contain very few parameters are omitted for clarity.}
    \begin{tabular}{c|ccccccc}
\toprule
 & ViT & Align module & Vision Resampler & $D_{tag}$ & $D_{rtc}$ & CLIP & LLM\tabularnewline
\midrule
Trainable & \ding{55} & \ding{51} & \ding{51} & \ding{51} & \ding{51} & \ding{55} & \ding{55}\tabularnewline
Parameters (\%) & 23.78 & 0.20 & 2.53 & 0.05 & 0.05 & 2.99 & 68.79\tabularnewline
Flops (G) & 783.5 & 2.1 & 6.4 & 6.2 & 0.4 & 6.5 & 80.1 \tabularnewline
\bottomrule
\end{tabular}
    \label{tab:supp_parameters}
\end{table*}

\section{Details of the Control Embeddings}

Following ControlCap~\cite{zhao2024controlcap}, we introduce control words to alleviate the caption degeneration issue, which refers to the fact that pre-trained multimodal models tend to predict the most frequent captions but miss the less frequent ones. During training, the control words are parsed from the ground-truth captions (Fig.~\ref{fig:supp_control_words}) and are randomly dropped in accordance with a Bernoulli distribution, which is detailed in Fig.~\ref{fig:supp_control_words}. The remaining control words are shuffled and combined with a $\texttt{[SEP]}$ token to form a control sentence, $i.e.$, ``$\texttt{white dog, sofa[SEP]}$'' in Fig.~\ref{fig:decoder}. The sentence is encoded into the control embedding by the tokenizer and word embedding layer of the large language model. During inference, we build the control embeddings with high-confidence tags from the outputs of \Ours.

\section{Illustration of pHASH operation}

The pHASH (Perceptual Hash) operation is a hashing algorithm that generates a "perceptual fingerprint" of an image based on its visual characteristics. 
The key features of pHASH operation are summaries as follows:

\textbf{i) Perceptual Similarity:} The pHASH operation is designed to generate similar hash values for visually similar images. It focuses on the aspects of the image that humans perceive ($e.g.$, shapes, colors).

\textbf{ii) Tolerance to Minor Modifications:} The pHASH operation is robust to minor changes like resizing, cropping, compression, or slight color variations. This tolerance makes it ideal for detecting duplicates or near-duplicates of images.

\textbf{iii) Fixed-Length Output:}
The output of the pHASH operation is always a fixed-length binary string (e.g., 64 or 128 bits), regardless of the size of the input image. This makes it easy to compare images of varying sizes.

\textbf{iv) Fast Computation:}
The pHASH operation is optimized for speed and is computationally efficient, allowing it to be used for large amount image comparisons.

\section{Detailed Experimental Settings}
\label{app:experiment}

\begin{table}[t]
    \small
    \centering
    \tabcolsep=0.15cm
    \caption{Detailed hyperparameters during training and inference.}
    \begin{tabular}{l | c}
\toprule
\textbf{Training} & \textbf{Value}\tabularnewline
\midrule
GPUs & 8$\times$ A800 80G\tabularnewline
batch size & 512\tabularnewline
training epochs & 5\tabularnewline
learning policy & cosine annealing\tabularnewline
initial learning rate & 1$e$-4\tabularnewline
minimum learning rate & 0\tabularnewline
weight decay ratio & 0.05\tabularnewline
warmup steps & 5000\tabularnewline
\midrule
\textbf{Inference} & \textbf{Value}\tabularnewline
\midrule
number of beams & 5 \tabularnewline
number of views & 3 (default) \tabularnewline
view selection & Image prior (default) \tabularnewline
\bottomrule
\end{tabular}
    \label{tab:supp_optimization}
\end{table}

The detailed model, dataset, evaluation settings of \Ours is summarized as follows:

\textbf{Model implementation.}  \Ours is implemented upon the LAVIS~\cite{li2022lavis} framework, where large language model and vision resampler are respectively initialized by FlanT5$_{\text{XL}}$~\cite{Chung2022Flan5} and Q-former~\cite{Li2023BLIP2}.
All the sampled views are resized to $224\times224$ resolution.
All models are trained using 8 NVIDIA A800 GPUs by 5 epochs, with the Adam optimizer where the batch size is set to 512. The total training time is less than 20 hours.
The initial learning rate is set to $1 \times 10^{-4}$ with a cosine learning rate decay. The detailed hyperparameters during training and inference are shown in Tab.~\ref{tab:supp_optimization}.
Considering that dense captioning requires the model to initially generate dense bounding-boxes, we utilize a GRiT~\cite{Wu2022GRIT} model trained on the VG to acquire object locations.
During the inference stage, we use the bounding boxes and object scores predicted by GRiT, and then replace its predicted caption with \Ours to get the final result.

\textbf{Datasets.} For all tasks, \Ours is trained using Visual Genome (VG)~\cite{krishna2017visual} and RefCOCOg~\cite{yu2016modeling}. For ablation studies, \Ours is trained using VG-COCO~\cite{Shao2022Region} and RefCOCOg~\cite{yu2016modeling}. For evaluation, we evaluate the region-level captioning performance on VG, VG-COCO~\cite{Shao2022Region}, and RefCOCOg, the open vocabulary attribute detection performance on OVAD~\cite{ovad}, and the region recognition performance on COCO~\cite{coco}

\textbf{Evaluation Metrics.} 
For region-level captioning, the METEOR score and CIDEr score are adopted as the evaluation metrics following ~\cite{glamm, osprey, guo2024regiongpt}.
For dense captioning, mean Average Precision (mAP)~\cite{Johnson2016DenseCap} is adopted as the evaluation metric following~\cite{Johnson2016DenseCap, Long2023CapDet}. The mAP is calculated across a range of thresholds for both localization and language accuracy, $i.e.$, the intersection over union (IoU) thresholds (0.3, 0.4, 0.5, 0.6, 0.7) are used for localization and the METEOR score’ thresholds (0, 0.05, 0.1, 0.15, 0.2, 0.25) is adopted for evaluating the language generation. Since \Ours lacks the capability to perform object detection, we utilize a GRiT~\cite{Wu2022GRIT} model trained on VG to acquire object locations. For open vocabulary attribute detection, mAP is adopted as the evaluation metric following OVAD~\cite{ovad}. For region recognition, mAP and Accuracy (Acc.) are are adopted as the evaluation metrics following ~\cite{guo2024regiongpt, regionclip}.

\section{Additional Experimental Results}



We provide additional experimental results in the supplementary as follows:

\textbf{Stochastic Multi-view Embedding: Align module.} The effectiveness of the align module is validated in Tab.~\ref{tab:abltion_align_module}. By spatially aligning the region embeddings across multiple views, \Ours achieves a 0.6\% improvement in mAP on OVAD, a 0.8\% improvement in mAP on VG-COCO, and a 5.1 increase in METEOR on RefCOCOg. These results validate the effectiveness of the proposed align module.

\textbf{Selectively Multimodal Referring.} 
As shown in Tab.~\ref{tab:abltion_task_prior}, we evaluate \Ours under different view counts and inference strategies. In the "No prior" strategy, views are randomly selected for each sample. In the "Task prior" strategy, the view containing the referred region is always selected, and the top-($n$-1) views are chosen based on the results from Fig.~4 for an $n$-view model. In the "Image prior" strategy, views are selected according to Eq.~1 in the main paper.
For the 2-view \Ours model, the performance of different strategies ranks as: ``Task prior > Image prior > No prior''. For the 3-view model, the ranking is: ``Task prior $\approx$ Image prior > No prior''. While the ``Task prior'' strategy works well, the "Image prior" strategy offers greater flexibility. It is task-independent and can dynamic select views to each image region. This makes it particularly suitable for models that need to handle multiple tasks with a unified region representation. Based on these advantages, we adopt ``Image prior'' as the default inference strategy.

\textbf{Statistics of Parameters and FLOPs.} The parameter and flop composition of \Ours is shown in Tab.~\ref{tab:supp_parameters}. \Ours has few trainable parameters and can be trained efficiently.

\begin{figure*}[ht]
	\centering\includegraphics[width=1.0\linewidth]{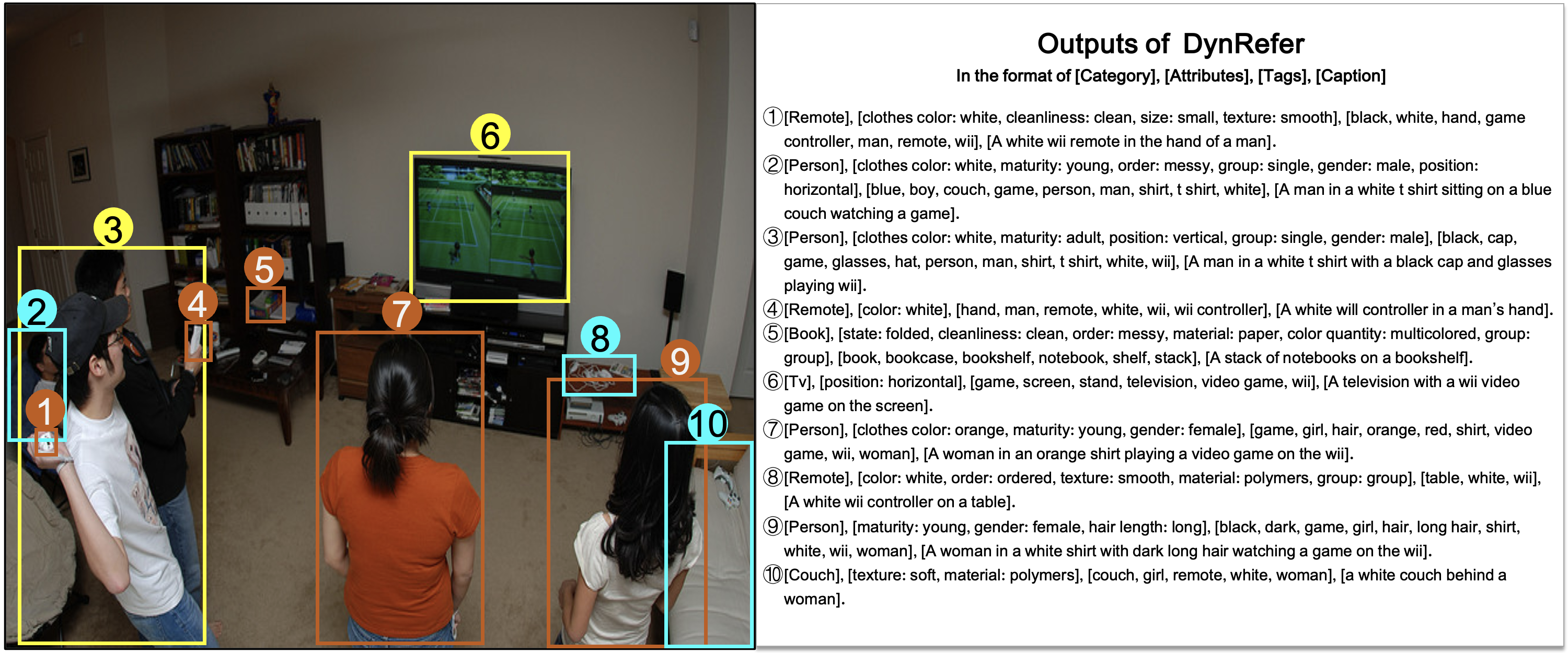}
     \caption{More results of Fig.~6 in the main paper, $i.e.$, illustration of DynRefer’s multi-task capability.}
    \label{fig:supp_demo4}
\end{figure*}

\begin{figure*}[ht]
	\centering\includegraphics[width=1.0\linewidth]{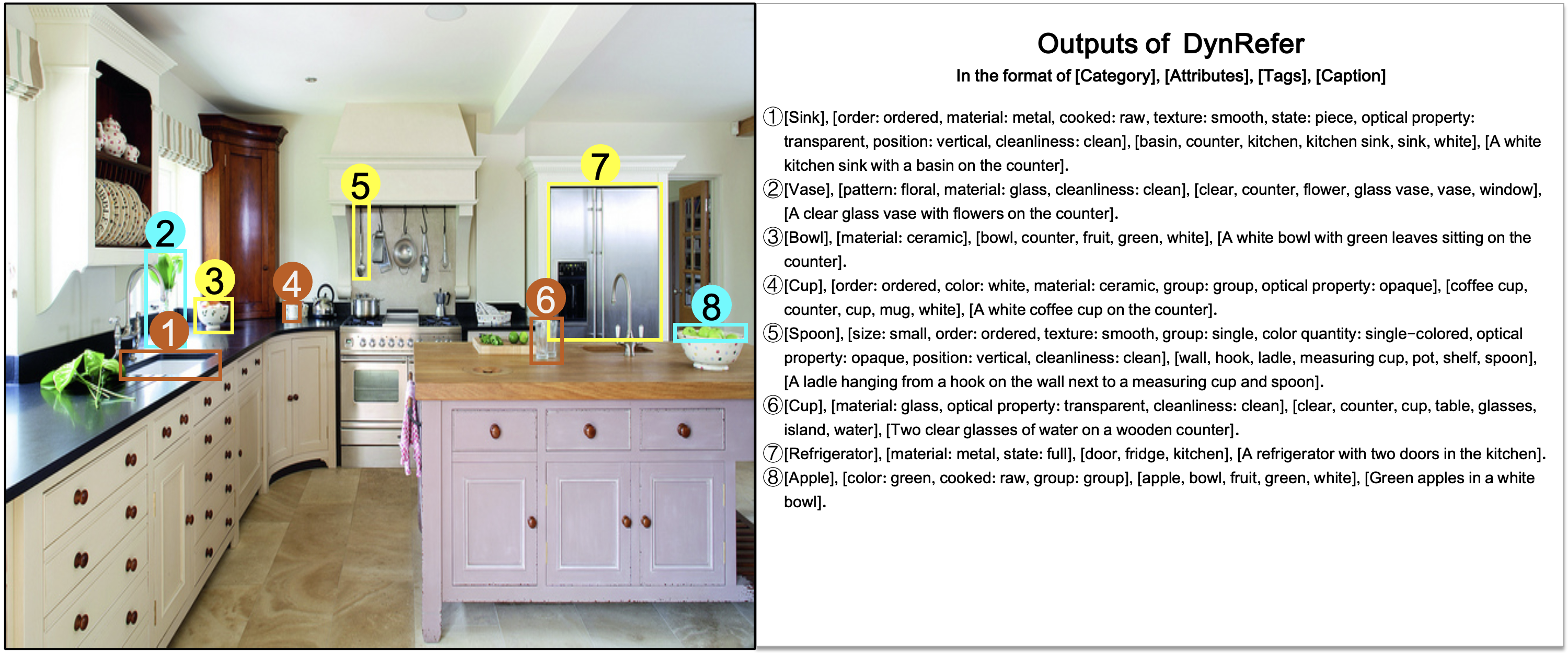}
     \caption{More results of Fig.~6 in the main paper, $i.e.$, illustration of DynRefer’s multi-task capability.}
    \label{fig:supp_demo5}
\end{figure*}

\begin{figure*}[ht]
	\centering\includegraphics[width=0.8\linewidth]{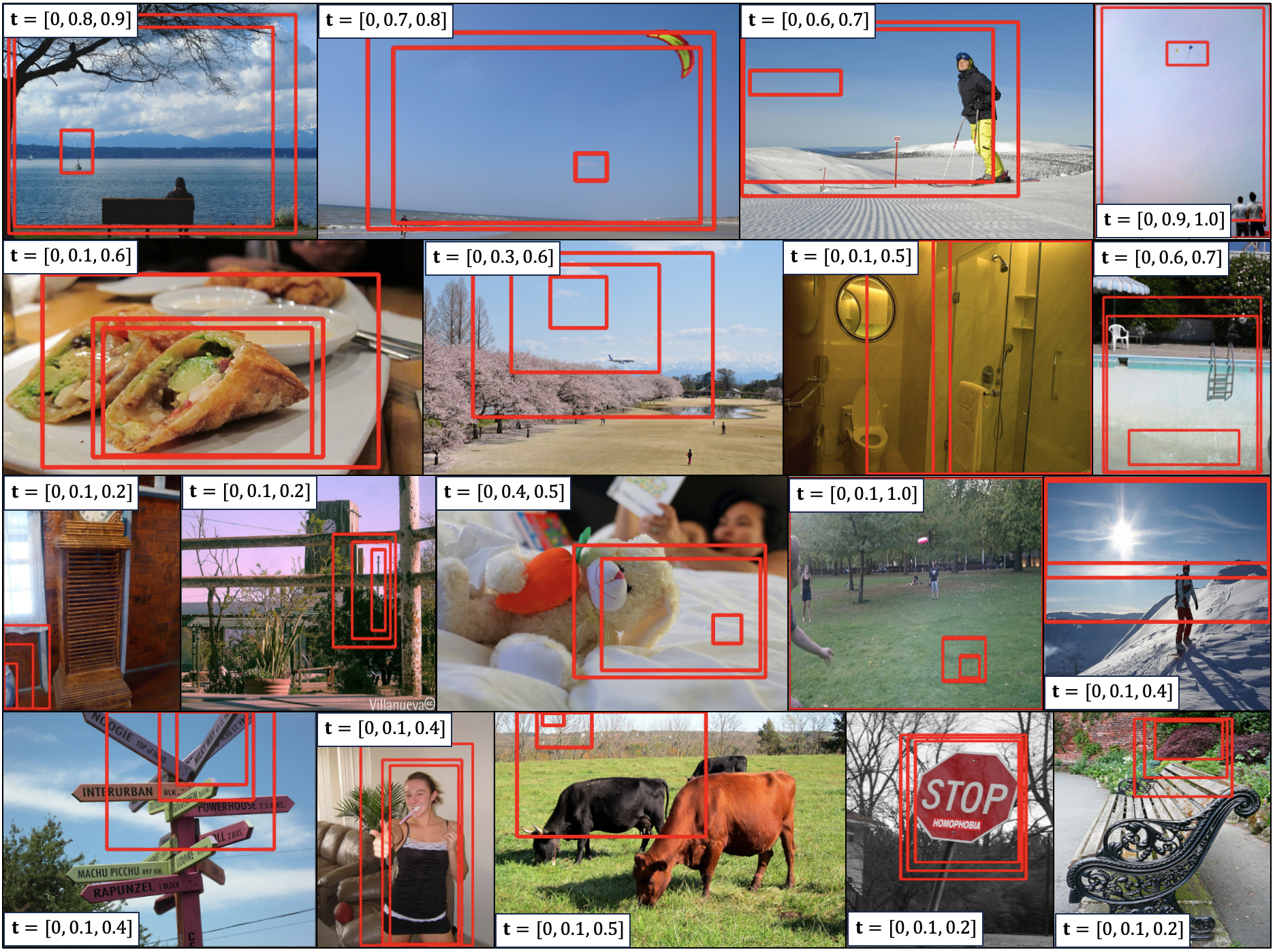}
     \caption{More results of Fig.~5 in the main paper, $i.e.$, visualization of selected views using image prior.}
    \label{fig:supp_image_prior}
\end{figure*}

\begin{figure*}[ht]
	\centering\includegraphics[width=0.75\linewidth]{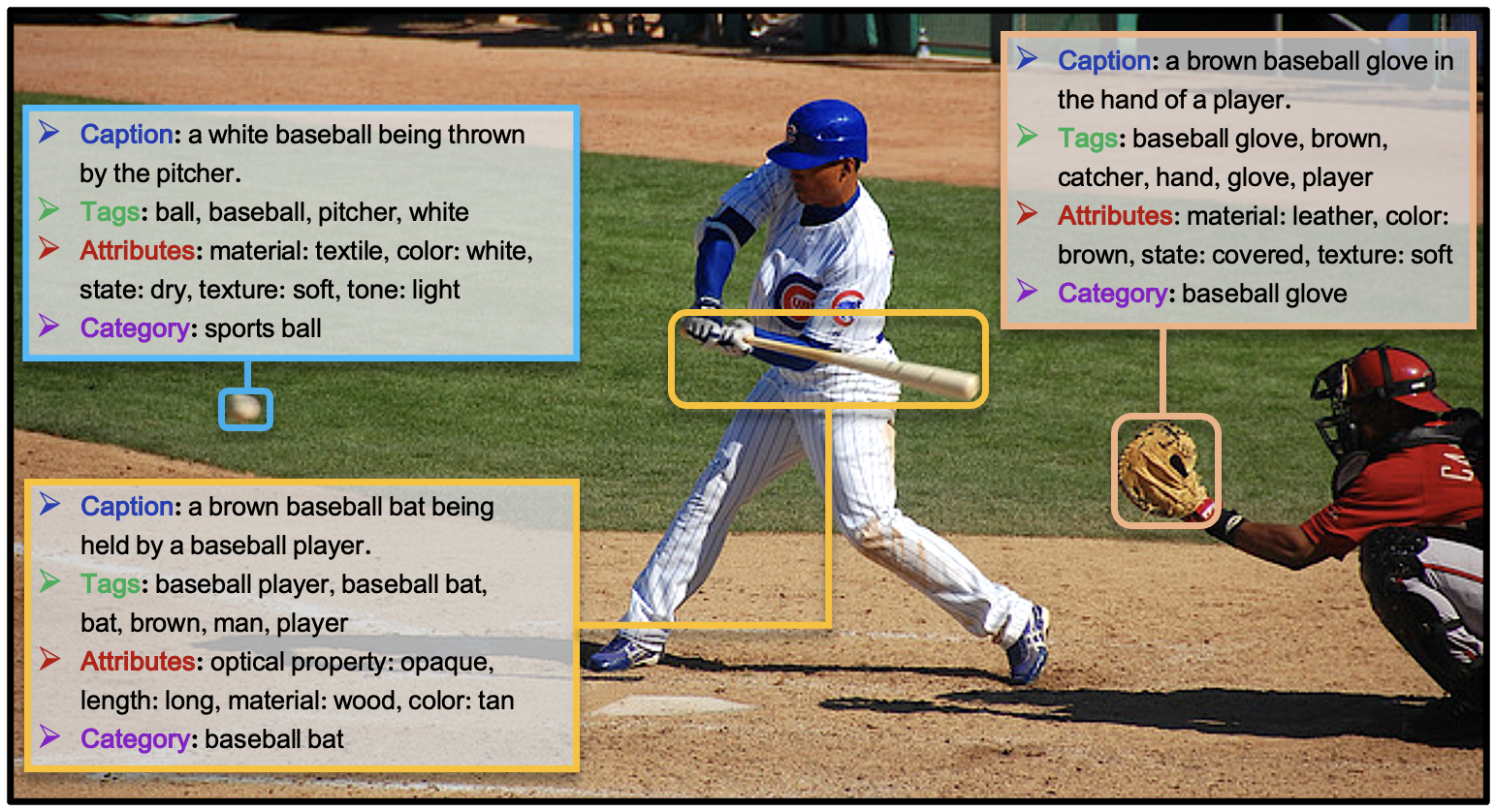}
     \caption{Illustration of \Ours's multi-task capability. It can generate captions, tags, attributes, categories, using a single model, for any referred regions.}
    \label{fig:supp_demo1}
\end{figure*}

\begin{figure*}[ht]
	\centering
\includegraphics[width=0.75\linewidth]{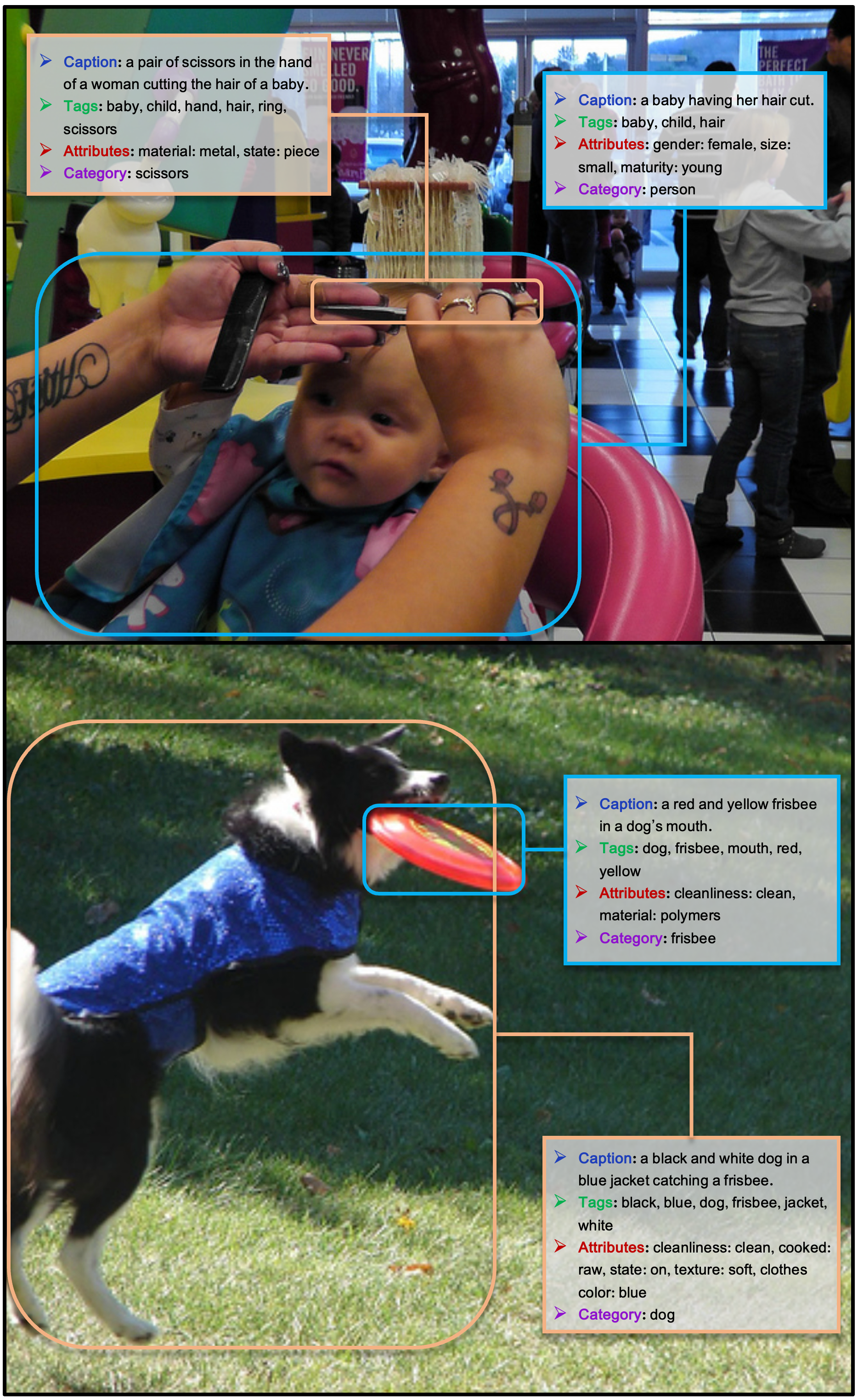}
     \caption{Illustration of \Ours's multi-task capability. It can generate captions, tags, attributes, categories, using a single model, for any referred regions.}
    \label{fig:supp_demo2}
\end{figure*}

\begin{figure*}[ht]
	\centering\includegraphics[width=0.75\linewidth]{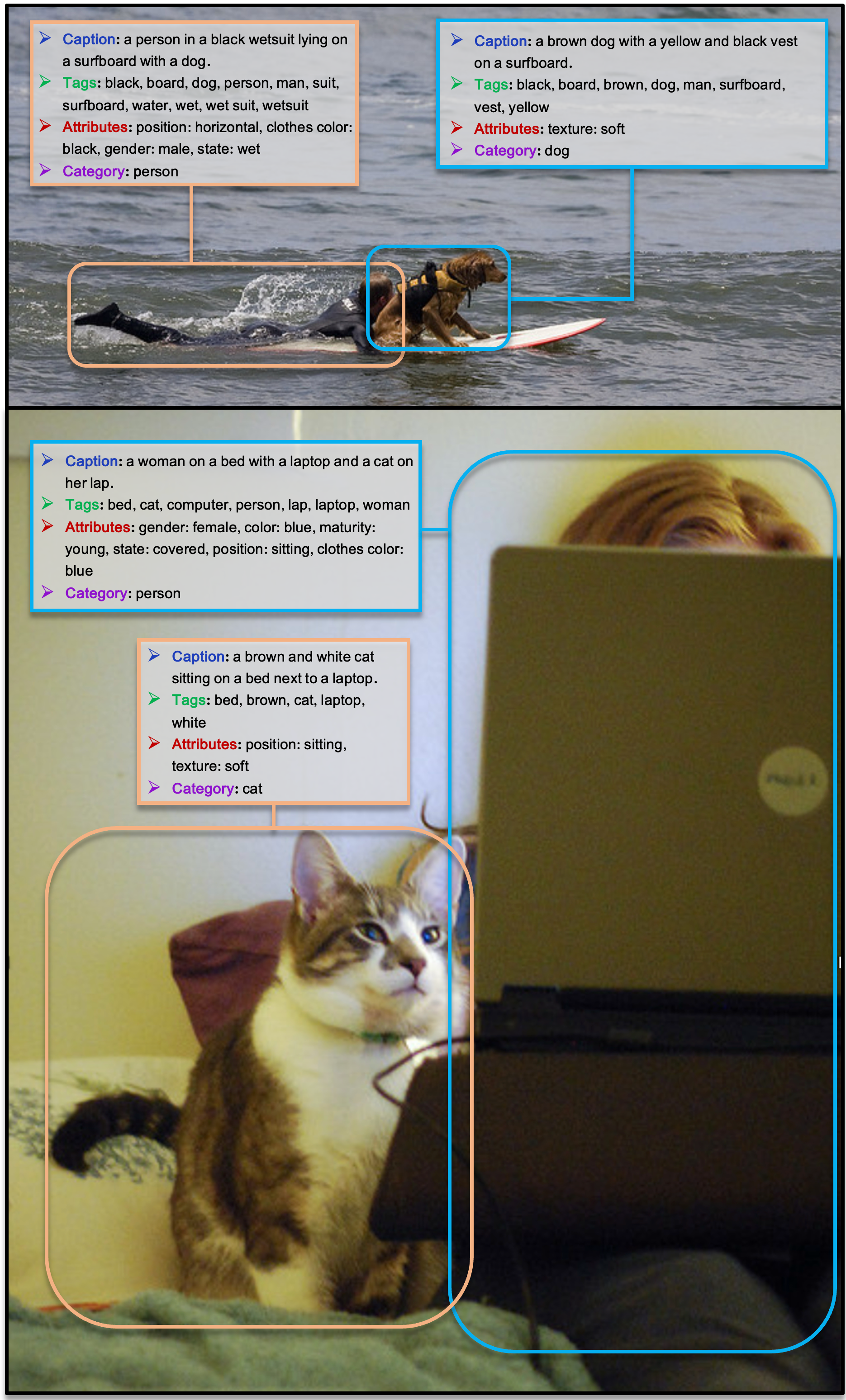}
     \caption{Illustration of \Ours's multi-task capability. It can generate captions, tags, attributes, categories, using a single model, for any referred regions.}
    \label{fig:supp_demo3}
\end{figure*}

\textbf{Additional Visualization Results.} We provide additional visualization results of Fig.~\ref{fig:demo_image_prior} and Fig.~\ref{fig:demo} in the main document. The results are shown in Fig.~\ref{fig:supp_image_prior} and Fig.~\ref{fig:supp_demo4}~\ref{fig:supp_demo5}~\ref{fig:supp_demo1}~\ref{fig:supp_demo2}~\ref{fig:supp_demo3}. 

\section{Limitations}
Though \Ours significantly outperforms previous state-of-the-arts on multiple multimodal tasks, it still doesn't perfectly mimic the visual cognition system of human. A real human can adjust the resolution of visual inputs in a more dynamic and flexible way. 
Better simulation strategy can be explored in the future work.


\end{document}